\documentclass[10pt,twocolumn,letterpaper]{article}

\usepackage{cvpr}              

\usepackage{graphicx}
\usepackage{amsmath}
\usepackage{amssymb}
\usepackage{booktabs}
\usepackage{multirow}

\usepackage{upgreek}
\usepackage{subcaption}

\usepackage{bigints}
\usepackage{bbm}

\usepackage{color}
\usepackage{colortbl}
\definecolor{secondbest}{rgb}{1.0,0.0,0.0}
\definecolor{gray}{rgb}{0.95,0.95,0.95}
\usepackage{hhline}
\usepackage[hang,flushmargin]{footmisc}
\usepackage{lipsum}

\newcommand\blfootnote[1]{%
	\begingroup
	\renewcommand\thefootnote{}\footnote{#1}%
	\addtocounter{footnote}{-1}%
	\endgroup
}

\usepackage[resetlabels]{multibib}
\newcommand{\bibnamec}{References}
\newcites{supp}{\bibnamec}
\usepackage[title]{appendix}
\usepackage[pagebackref,breaklinks,colorlinks]{hyperref}

\usepackage[capitalize]{cleveref}
\crefname{section}{Sec.}{Secs.}
\Crefname{section}{Section}{Sections}
\Crefname{table}{Table}{Tables}
\crefname{table}{Tab.}{Tabs.}

\def\cvprPaperID{1302} 
\def\confName{CVPR}
\def\confYear{2022}

\begin{document}

\title{Joint Distribution Matters: Deep Brownian Distance Covariance for \\Few-Shot  Classification}

\author{Jiangtao Xie$^{1, *}$, Fei Long$^{1, *}$, Jiaming Lv$^{1}$, Qilong Wang$^{2}$, Peihua Li$^{1, \dagger}$\\
$^{1}$Dalian University of Technology, China$\;\;$ $^{2}$Tianjin University, China\\
}
\maketitle

\begin{abstract}

Few-shot classification is a  challenging problem as only  very few  training examples are given  for each new task. One of  the effective research lines to address this challenge  focuses on learning deep representations driven by a similarity measure between a query image and few support images of some class. Statistically, this amounts to measure the dependency of image features, viewed as random vectors in a high-dimensional embedding space. Previous methods either  only use marginal distributions without  considering  joint distributions, suffering from limited representation capability, or  are computationally expensive though harnessing joint distributions. In this paper, we propose a deep Brownian Distance Covariance (DeepBDC) method for few-shot classification. The central  idea of DeepBDC is to learn image representations by measuring the  discrepancy between  joint characteristic functions of embedded features and  product of the marginals.  As the BDC metric is decoupled, we  formulate it  as a highly modular and efficient  layer. Furthermore, we instantiate DeepBDC in two different few-shot classification frameworks.    We make  experiments on six standard few-shot image benchmarks, covering general object recognition, fine-grained categorization and cross-domain classification. Extensive evaluations show  our DeepBDC  significantly outperforms the counterparts, while establishing  new state-of-the-art  results. The source code  is available at \href{http://www.peihuali.org/DeepBDC}{http://www.peihuali.org/DeepBDC}.

\end{abstract}

\section{Introduction}\label{sec:intro}

Few-shot  classification~\cite{One-shot-CogSci-2011,Koch2015SiameseNN} is concerned with a task where a classifier can be adapted to distinguish   classes unseen previously, given only a very limited number of examples of these classes\blfootnote{$^{*}$Equal contribution. $^{\dagger}$Corresponding author, peihuali@dlut.edu.cn. The work was supported by National Natural Science Foundation of China (61971086,
	 61806140), and CCF-Baidu Open Fund (2021PP15002000).}. This is a challenging problem as scarcely labeled examples are far from sufficient for learning abundant knowledge and also likely lead to overfitting. One practical solution is based on the technique of meta-learning or learning to learn~\cite{vinyals2016matching,finn2017model},  in which the episodic training is formulated to transfer the knowledge obtained on a massive meta-training set spanning a large number of known classes  to the few-shot regime of novel classes. Among great advances that have been made,  the line of metric-based methods  attracts considerable research interest~\cite{Koch2015SiameseNN,vinyals2016matching,snell2017prototypical,oreshkin2018tadam}, achieving state-of-the-art performance~\cite{zhang2020deepemd, Wertheimer_2021_CVPR} in recent years.

The primary idea of the metric-based few-shot classification is to learn representations through deep networks, driven by the similarity  measures between a query image  and few support images of some class~\cite{snell2017prototypical,zhang2020deepemd}. Statistically,  the features of a query image (resp., support images) can be viewed as observations of a random vector $X$ (resp., $Y$) in a high-dimensional embedding space. Therefore, the similarity between images can be measured by means of  probability distributions. However, modeling distributions of high-dimensional (and often few) features is hard and a common method is to model statistical moments.  ProtoNet~\cite{snell2017prototypical} and its variants (e.g.,~\cite{oreshkin2018tadam}) represent images by  first moment (mean vector)  and use  Euclidean distance or cosine similarity for metric learning. To capture richer statistics, several works study second moment (covariance matrix)~\cite{Wertheimer2019} or combination of first and second moments in the form of Gaussians~\cite{ADM} for image representations,  while adopting Frobenius norm or Kullback-Leiberler (KL) divergence as similarity measures. However, these methods only exploit marginal distributions while neglecting  joint distributions, limiting the performance of learned models. In addition, the covariances can only model linear relations.

\begin{table*}
	\centering
	\setlength{\tabcolsep}{3.5pt}
	\renewcommand\arraystretch{1.0}
	\footnotesize
	\begin{tabular}{c|ll|c|c|c|c|c}
		\hline
		\multirow{2}*{Method}& \multirow{2}*{\parbox{10mm}{Probability \\model}} & \multirow{2}*{Dis-similarity$/$similarity measure}& Joint & \multirow{2}*{Dependency} & \multirow{2}*{$\;\;\;$Latency$\;\;\;$} & \multicolumn{2}{c}{Accuracy (\%)}  \\
		\cline{7-8}
		&  && distribution & &  & 1-shot & 5-shot \\
		\hline
		ProtoNet~\cite{snell2017prototypical} & \parbox{10mm}{\vspace{1mm} Mean\\ vector\vspace{1mm}} & $\|\boldsymbol{\upmu}_{X}-\boldsymbol{\upmu}_{Y}\|^{2}$ $\,$or$\,$   $\frac{\boldsymbol{\upmu}_{X}^{T}\boldsymbol{\upmu}_{Y}}{\|\boldsymbol{\upmu}_{X}\|\|\boldsymbol{\upmu}_{Y}\|}$ & No & N/A & Low & 49.42 & 68.20 \\
		\hline
		CovNet~\cite{Wertheimer2019} &\parbox{15mm}{\vspace{1mm}Covariance\\ matrix\vspace{1mm}} & $\|\boldsymbol{\Sigma}_{X}-\boldsymbol{\Sigma}_{Y}\|^{2}$& No & Linear & Low & 49.64 & 69.45 \\
		\hline
		ADM~\cite{ADM} & \parbox{10mm}{\vspace{1mm}Gaussian\\ distribution\vspace{1mm}} & \parbox{35mm}{\vspace{1mm}\centering  $D_{\mathrm{KL}}(\mathcal{N}_{\boldsymbol{\upmu}_{X},\boldsymbol{\Sigma}_{X}}||\mathcal{N}_{\boldsymbol{\upmu}_{Y},\boldsymbol{\Sigma}_{Y}})$} & No & N/A & Low & 53.10 & 69.73 \\
		\hline
		DeepEMD~\cite{zhang2020deepemd} & \parbox{10mm}{\vspace{1mm}Discrete \\distribution\vspace{1mm}} & \parbox{55mm}{\vspace{1mm}  
			$\min_{f_{\mathbf{x}_j,\mathbf{y}_l}\geq 0}\sum_{j}\!\!\sum_{l}f_{\mathbf{x}_j,\mathbf{y}_l}c_{\mathbf{x}_j,\mathbf{y}_l}$\\ $\mathrm{s.t.}\; \sum_{l}\!f_{\mathbf{x}_j,\mathbf{y}_l}\!\!=\!\!f_{\mathbf{x}_{j}},\sum_{j}\!f_{\mathbf{x}_j,\mathbf{y}_l}\!\!=\!\!f_{\mathbf{y}_{l}}$ for $\forall j,l$
		} & Yes & N/A & High & 65.91 & 82.41 \\
		\hline
		\hline
		DeepBDC (ours) &\parbox{10mm}{\vspace{1mm} Characteristic\\ function\vspace{1mm}}& \parbox{55mm}{ $\bigintsss_{\mathbb{R}^{p}}\!\!\!\bigintsss_{\mathbb{R}^{q}}\!\!\!\dfrac{|\phi_{XY}(\mathbf{t},\mathbf{s})\!-\!\phi_{X}(\mathbf{t})\phi_{Y}(\mathbf{s})|^2}{c_{p}c_{q}\|\mathbf{t}\|^{1+p}\|\mathbf{s}\|^{1+q}}d\mathbf{t}d\mathbf{s}$} & Yes & \parbox{18mm}{\centering Nonlinear \&\\ Independence} & Low & \textbf{67.34} & \textbf{84.46} \\
		\hline
		
	\end{tabular}
	\caption{Comparison between our DeepBDC and the counterparts. To quantify the dependency between  random vectors $X$ and $Y$,  moments based methods~\cite{snell2017prototypical,Wertheimer2019,ADM} only model marginal distributions, suffering from limited representation capability; though achieving state-of-the-art performance by considering joint distributions,  DeepEMD~\cite{zhang2020deepemd} is  computationally  expensive. Our DeepBDC measures discrepancy between  joint characteristic function and  product of the marginals, which can be efficiently computed in closed-form, and model non-linear relations and fully characterizes independence. Note that for a random vector  its characteristic function and probability distribution are equivalent in that they form a Fourier transform  pair. Here we report  accuracies of 5-way 1-shot$/$5-shot classification on $mini$ImageNet;  our result is obtained by Meta DeepBDC and results of the counterparts are duplicated from respective papers.}	
	\label{tab:BDC vs. previous methods} 
	\vspace{-12pt}
\end{table*}

 In general, the dependency between $X$ and $Y$ should  be measured in light of  their joint distribution $f_{XY}(\mathbf{x},\mathbf{y})$~\cite{Book-Information}.   Earth Mover's Distance (EMD) is an effective method for measuring such dependency. As described in~\cite[Sec. 2.3]{MAL-073}, EMD seeks an optimal joint distribution $f_{XY}(\mathbf{x},\mathbf{y})$, whose marginals are constrained to be given $f_{X}(\mathbf{x)}$ and $f_{Y}(\mathbf{y})$, so that the expectation of transportation cost is minimal.
In few-shot classification, DeepEMD~\cite{zhang2020deepemd} proposes  differential EMD for optimal matching of image regions. Though achieving state-of-the-art performance, DeepEMD is computationally  expensive~\cite{Wertheimer_2021_CVPR}, due to  inherent linear programming algorithm. 
Mutual information (MI)~\cite{Book-PRML,Mutual_information_PAMI} is a well-known measure, which can quantify the dependency of two random variables by  KL-divergence between their joint distribution and  product of the marginals. Unfortunately, computation of MI is  difficult in  real-valued, high-dimensional setting~\cite{MINE}, and often involves difficult density modeling or lower-bound estimation of KL-divergence~\cite{DeepDIM}.

In this paper, we propose a deep Brownian Distance Covariance (DeepBDC) method for few-shot  classification.  The BDC metric, first proposed in~\cite{Szekely2007,Szekely2009},  is defined as the Euclidean distance between the joint characteristic function and  product of the marginals. It can naturally quantify the dependency between two random variables.  For discrete observations (features), the BDC metric is decoupled so that  we can formulate BDC as a pooling layer, which can be seamlessly inserted into a deep network, accepting feature maps as input and outputting a BDC matrix as an image representation. In this way, the similarity between  two images is  computed as the inner product between the corresponding two BDC matrices. Therefore, the  core of  our DeepBDC  is highly modular and  plug-and-play for different methodologies of few-shot image classification. Specifically, we instantiate our DeepBDC in meta-learning framework (Meta DeepBDC), and in the simple transfer learning framework relying non-episodic training (STL DeepBDC).  Contrary to covariance matrices,  our DeepBDC can freely handle non-linear relations and fully characterize independence. Compared to EMD, it also considers joint distribution and above all, can be computed analytically and efficiently. Unlike MI, the BDC requires no density modeling.  We present differences between  our BDC and the counterparts in Tab.~\ref{tab:BDC vs. previous methods}.

Our contributions are summarized as follows. (1)  For the first time, we introduce Brownian distance covariance (BDC), a fundamental but largely overlooked dependency modeling method, into deep  network-based  few-shot classification. Our work suggests great potential  and  future applications of BDC in deep learning. (2)  We formulate DeepBDC as a highly modular and efficient layer, suitable for different few-shot learning frameworks. Furthermore, we propose two instantiations for few-shot classification, i.e., Meta DeepBDC based on the meta-learning framework with ProtoNet as a blue print,  and STL DeepBDC based on simple transfer learning framework without episodic training.  (3)  We perform thorough ablation study on our methods and conduct extensive experiments on six few-shot classification benchmarks.  The experimental results  demonstrate that  both of our two instantiations achieve superior performance and meanwhile set  new  state-of-the-arts.

\section{Related Works}

\vspace{3pt}\noindent \textbf{Representation learning in few-shot classification}  The image representation and the similarity measure  play  important roles in few-shot classifications where only limited labeled examples are available. In light of the image representation, we can roughly divide the few-shot classification methods into two categories. In the first category, the image representations are based on distribution modeling. They use  either first moment (mean vector)~\cite{snell2017prototypical}, second moment (covariance matrix)~\cite{Wertheimer2019} , Gaussian distribution~\cite{ADM} or discrete probability~\cite{zhang2020deepemd}, and, accordingly, adopts Euclidean distance (or cosine similarity), Frobenious norm,  KL-divergence or Earther Mover's Distance as dis-similarity measures. The second category is concerned with  feature  reconstruction between the query image and the support images,  by means of either  directly linear reconstruction through  Ridge regression~\cite{Wertheimer_2021_CVPR} or attention mechanism~\cite{CTX,ye2020few}, or concerned with designing  relational module to learn a transferable  deep metric~\cite{sung2018learning,zhang2019power}. Our methods belong to the first category, and the biggest difference from existing works is that we use Brownian Distance Covariance for representation learning in the few-shot regime.

\vspace{3pt}\noindent \textbf{Meta-learning versus simple transfer learning}   Meta-learning is  a de facto  framework for few-shot classification~\cite{vinyals2016matching,finn2017model}. It involves a family of tasks (episodes) split into disjoint meta-training and meta-testing sets. Typically, each task is formulated as a $N$-way $K$-shot classification, which spans $N$ classes each provided with $K$ support images and some query images. The meta-training and meta-testing sets share the episodic training strategy that facilitates  generalization ability across tasks.  Most of the  methods, either optimization-based~\cite{finn2017model,DBLP:journals/corr/abs-1909-09157} or  metric-based~\cite{snell2017prototypical,sung2018learning},  follow this methodology. A lot of studies~\cite{zhang2020deepemd,Wertheimer_2021_CVPR,chen2021meta} have shown that, rather than meta-training from scratch, pre-training on the whole meta-training set is  helpful for meta-learning.  Recently, it has been found that  simple transfer learning (STL) framework, which does not rely on episodic training at all, achieves very competitive performance~\cite{chen2019closer,baseline-iclr-2020,tian2020rethinking}. For STL methods, during meta-training a deep network is trained for a common classification problem via standard cross-entropy loss on the whole meta-training set spanning all  classes;   during meta-testing, the trained model is used as an embedding model for feature extraction, then a linear model, such as a soft-max classifier~\cite{chen2019closer,baseline-iclr-2020} or logistic regression model~\cite{tian2020rethinking}, is constructed and trained for the few-shot classification. 

Finally, we mention that scarce works have ever used BDC in machine learning or computer vision, and so far we  find one  BDC-based dimension reduction method~\cite{pmlr-v54-cowley17a}  which is not concerned with deep learning.

\section{Proposed Method}\label{section: proposed method}

In this section, we first  introduce Brownian distance covariance (BDC). Then we formulate our DeepBDC in the convolutional networks. Finally, we instantiate our DeepBDC for few-shot image classification.  

\subsection{Brownian Distance Covariance (BDC)}\label{subsection:BDC theory}

The theory of  BDC  is first established in~\cite{Szekely2007,Szekely2009} in light of  characteristic function. The characteristic function of a random vector is equivalent to its probability density function (PDF), as they form a Fourier transform pair. 

 Let $X \in \mathbb{R}^{p}, Y \in \mathbb{R}^{q}$ be random vectors of dimension $p$ and $q$, respectively, and let  $f_{XY}(\mathbf{x},\mathbf{y})$  be their joint PDF.  The joint characteristic function of $X$ and $Y$ is defined as 
\begin{align}
\!\!\!\!\phi_{XY}(\mathbf{t},\mathbf{s})\!=\!\!\!
\int_{\mathbb{R}^{p}}\!\!\int_{\mathbb{R}^{q}}\!\!\exp(\!i(\mathbf{t}^{T}\mathbf{x}\!+\!\mathbf{s}^{T}\mathbf{y}))f_{XY}(\mathbf{x},\mathbf{y})d\mathbf{x}d\mathbf{y}
\end{align}
where $i$ is the imaginary unit. Clearly, the marginal distributions of $X$ and $Y$ are respectively $\phi_{X}(\mathbf{t})\!\!=\!\!\phi_{XY}(\mathbf{t},\mathbf{0})$ and  $\phi_{Y}(\mathbf{s})\!\!=\!\!\phi_{XY}(\mathbf{0},\mathbf{s})$ where $\mathbf{0}$ is a vector whose elements are all zero. From theory of probability, we know  $X$ and $Y$ are independent if and only if  $\phi_{XY}(\mathbf{t},\mathbf{s})\!\!=\!\!\phi_{X}(\mathbf{t})\phi_{Y}(\mathbf{s})$.  Provided $X$ and $Y$  have  finite first moments,  the BDC metric is defined as
\begin{align}\label{equ:definition BDC}
\rho(X,Y)\!=\!\int_{\mathbb{R}^{p}}\!\!\int_{\mathbb{R}^{q}}\!\!\dfrac{|\phi_{XY}(\mathbf{t},\mathbf{s})-\phi_{X}(\mathbf{t})\phi_{Y}(\mathbf{s})|^2}{c_{p}c_{q}\|\mathbf{t}\|^{1+p}\|\mathbf{s}\|^{1+q}}d\mathbf{t}d\mathbf{s}
\end{align}
where $\|\cdot\|$ denotes  Euclidean norm,  $c_{p}=\pi^{(1+p)/2}/\Gamma((1+p)/2)$ and $\Gamma$ is the complete gamma function. 

For the set of $m$  observations $\{(\mathbf{x}_{1},\mathbf{y}_{1}),\ldots, (\mathbf{x}_{m},\mathbf{y}_{m})\}$ which are independent and identically distributed  (i.i.d.), a natural approach is to define the BDC metric in light of the empirical characteristic functions:
\begin{align}
\phi_{XY}(\mathbf{t},\mathbf{s})=\frac{1}{m}\sum_{k=1}^{m}\exp(i(\mathbf{t}^{T}\mathbf{x}_{k}+\mathbf{s}^{T}\mathbf{y}_{k}))
\end{align}
Though Eq.~(\ref{equ:definition BDC}) seems complicated, the BDC metric has a closed form expression for discrete observations. 
Let $\widehat{\mathbf{A}}\!=\!(\widehat{a}_{kl})\!\in\! \mathbb{R}^{m\times m}$ where  $\widehat{a}_{kl}\!=\!\|\mathbf{x}_{k}\!-\!\mathbf{x}_{l}\|$ be an  Euclidean distance matrix computed between the pairs of  observations of $X$. Similarly, we compute the Euclidean distance  matrix $\widehat{\mathbf{B}}\!=\!(\widehat{b}_{kl}) \!\in\! \mathbb{R}^{m\times m}$ where  $\widehat{b}_{kl}\!=\!\|\mathbf{y}_{k}\!-\!\mathbf{y}_{l}\|$. Then the  BDC metric has the following form~\cite{Szekely2009}
\footnote{Actually, $\rho(X,Y)=\frac{1}{m^2}\mathrm{tr}\big(\mathbf{A}^{T}\mathbf{B}\big)$ and the constant $\frac{1}{m^2}$ is assimilated into a learnable scaling parameter $\tau$ (see Sec.~\ref{subsection:instantiation}) and thus is left out.}:
\begin{align}\label{equ: BDC measure}
\rho(X,Y)=\mathrm{tr}\big(\mathbf{A}^{T}\mathbf{B}\big)
\end{align}
where $\mathrm{tr}(\cdot)$ denotes matrix trace, $T$ denotes matrix transpose,  and  $\mathbf{A}=(a_{kl})$ is dubbed \textit{BDC matrix}. Here    $a_{kl}=\widehat{a}_{kl}-\frac{1}{m}\sum_{k=1}^{m}\widehat{a}_{kl}-\frac{1}{m}\sum_{l=1}^{m}\widehat{a}_{kl}-\frac{1}{m^{2}}\sum_{k=1}^{m}\sum_{l=1}^{m}\widehat{a}_{kl}$, where the last three terms indicate  means of the $l$-th column,  $k$-th row and  all entries of $\widehat{\mathbf{A}}$, respectively.  The matrix $\mathbf{B}$ can be computed in a similar manner from $\widehat{\mathbf{B}}$.  As a BDC matrix is symmetric, $\rho(X,Y)$ can also be written as the inner product of two BDC vectors $\mathbf{a}$ and $\mathbf{b}$, i.e., 
\begin{align}\label{equ: BDC measure-vector}
\rho(X,Y)=\big<\mathbf{a},\mathbf{b}\big>=\mathbf{a}^{T}\mathbf{b}
\end{align}
where  $\mathbf{a}$ (resp., $\mathbf{b}$) is obtained by extracting the upper triangular portion of $\mathbf{A}$ (resp., $\mathbf{B}$) and then performing vectorization.

The metric $\rho(X,Y)$  has some desirable properties. (1) It is non-negative,  and is equal to 0 if and only if $X$ and $Y$ are independent. (2) It can characterize linear  and non-linear dependency between  $X$ and $Y$. (3) It  is invariant to individual translations and orthonormal transformations of $X$ and $Y$, and equivariant to their individual scaling factors. That is, for any  vectors $\mathbf{c}_{1}\in \mathbb{R}^{p}, \mathbf{c}_{2}\in \mathbb{R}^{q}$, scalars $s_{1}, s_{2}$ and orthonormal matrices $\mathbf{R}_{1}\in \mathbb{R}^{p\times p}, \mathbf{R}_{2}\in \mathbb{R}^{q\times q}$, $\rho(\mathbf{c}_{1}+s_{1}\mathbf{R}_{1}X, \mathbf{c}_{2}+s_{2}\mathbf{R}_{2}Y)=|s_{1}s_{2}|\rho(X,Y)$.

\subsection{Formulation of DeepBDC  as a Pooling Layer}\label{subsection:BDC-computation}

In terms of Eq.~(\ref{equ: BDC measure}) and Eq.~(\ref{equ: BDC measure-vector}), we can see that the BDC metric is decoupled in the sense that we can independently compute the BDC matrix for every input image. Specifically, we design   a two-layer  module suitable for a convolutional network, which performs dimension  reduction  and computation of  the BDC matrix, respectively.  As the size of a BDC matrix increases quadratically with respect to  the number of channels (feature maps) in the network, we insert a $1\times 1$  convolution layer for  dimension reduction  right after the last convolution  layer of the network backbone.

Suppose the network (including the dimension reduction layer) is  parameterized by $\boldsymbol{\uptheta}$ which embeds a color image $\mathbf{z}\in \mathbb{R}^{3}$ into a feature space.  The  embedding of the image is a  $h \times w \times d$ tensor, where $h$ and $w$ are spatial height and width while $d$ is the number of channels. We reshape the tensor to a matrix $\mathbf{X}\in \mathbb{R}^{hw \times d}$, and can view either each column $\boldsymbol{\chi}_{k}\in\mathbb{R}^{hw}$ or  each row (after transpose)  $\mathbf{x}_{j}\in\mathbb{R}^{d}$ as an observation of random vector $X$.  We mention that in practice,  for either case the i.i.d. assumption may not hold, and comparison of the two options is given in  \hyperref[i.i.d.]{Sec. 4.2}. 

In what follows, we take for example  $\boldsymbol{\chi}_{k}$  as a random observation.   We develop three operators, which successively compute the squared Euclidean distance matrix $\widetilde{\mathbf{A}}=(\tilde{a}_{kl})$ where $\tilde{a}_{kl}$ is  squared Euclidean distance between the $k$-th column and $l$-th column of $\mathbf{X}$, the Euclidean distance matrix $\widehat{\mathbf{A}}=(\sqrt{\widetilde{a}_{kl}})$, and the  BDC matrix $\mathbf{A}$ obtained by subtracting from $\widehat{\mathbf{A}}$ its row mean, column mean and   mean of all of its elements. That is, 
\begin{align}\label{equ:computaton of BDC matrix}
\widetilde{\mathbf{A}}&=2\big(\mathbf{1}(\mathbf{X}^{T}\mathbf{X}\circ \mathbf{I})\big)_{\mathrm{sym}}-2\mathbf{X}^{T}\mathbf{X}\\
\widehat{\mathbf{A}}&=\big(\sqrt{\widetilde{a}_{kl}}\big)\nonumber\\
\mathbf{A}&=\widehat{\mathbf{A}}-\frac{2}{d}\big(\mathbf{1}\widehat{\mathbf{A}}\big)_{\mathrm{sym}}+\frac{1}{d^{2}}\mathbf{1}\widehat{\mathbf{A}}\mathbf{1}\nonumber
\end{align}
Here $\mathbf{1}\in \mathbb{R}^{d\times d}$ is a matrix each element of which is 1, $\mathbf{I}$ is the identity matrix, and $\circ$ indicates the Hadamard product. We denote $(\mathbf{U})_{\mathrm{sym}}=\frac{1}{2}(\mathbf{U}+\mathbf{U}^{T})$. Hereafter, we use $\mathbf{A}_{\boldsymbol{\uptheta}}(\mathbf{z})$ to indicate that the BDC matrix is computed from the network parameterized by $\boldsymbol{\uptheta}$ with an input image $\mathbf{z}$.

As such, we formulate  DeepBDC as  a parameter-free, spatial pooling layer. It is highly modular, suitable for varying network   architectures and  for different frameworks of few-shot classification.    The BDC matrix mainly involves standard matrix operations,  appropriate for parallel computation on GPU. From Eq.~(\ref{equ:computaton of BDC matrix}), it is clear that  the BDC matrix models non-linear relations among  channels through Euclidean distance. The covariance matrices can be interpreted similarly, which, however, models linear relations among channels through inner product~\cite[Sec. 4.1]{Beyond_IJCV2021}.  Theoretically, they are quite different as the BDC matrices consider the joint distributions, while the covariance matrices only consider the marginal ones. 

\begin{figure}[t!]
	\centering
	\begin{subfigure}[b]{0.5\textwidth}
		\centering
		\includegraphics[height=2.1in]{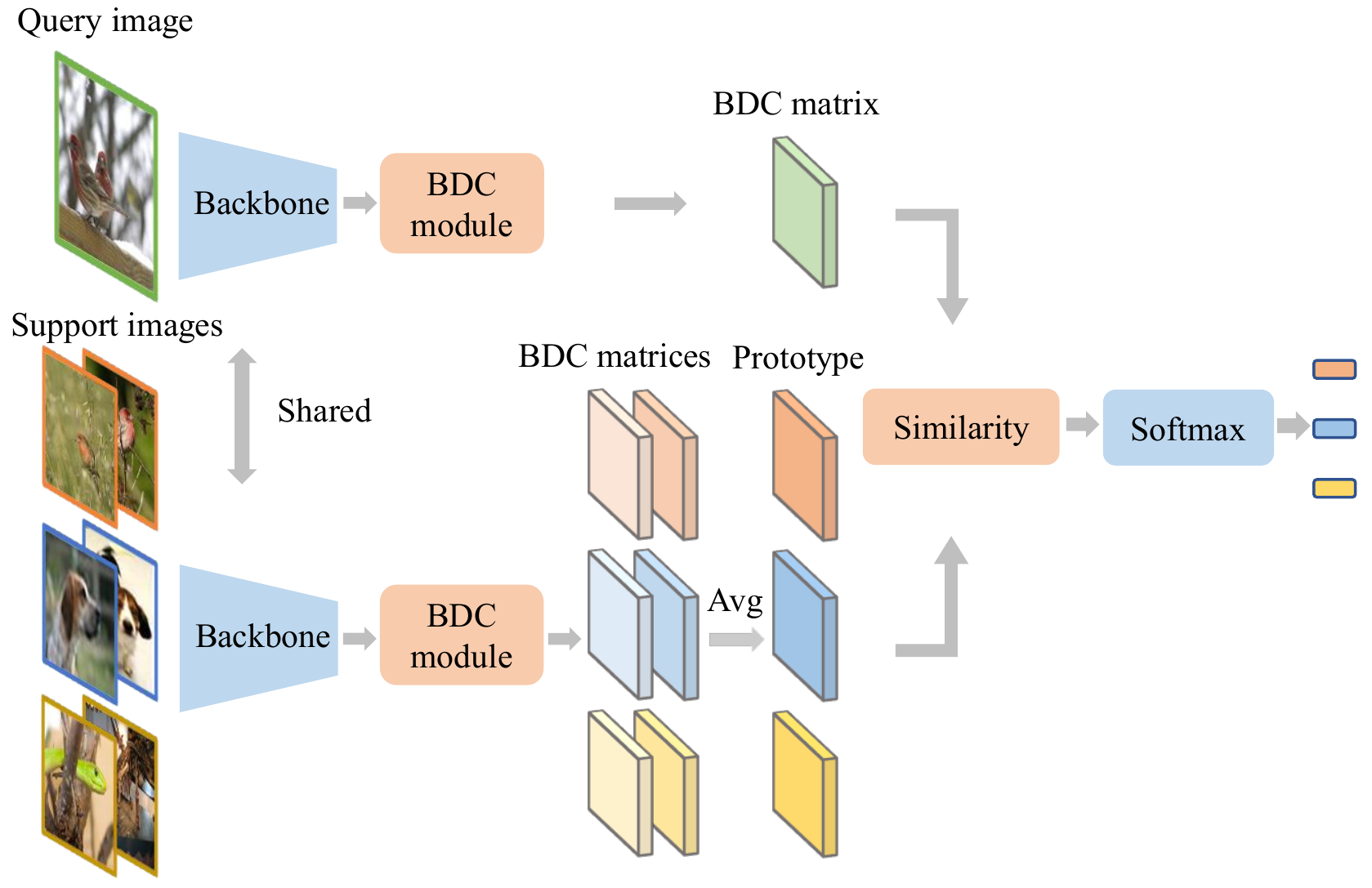}
		\caption{Meta DeepBDC--Instantiation with ProtoNet~\cite{snell2017prototypical} as a blueprint.}
		\label{subfigure:overveiw_a}
		\vspace{6pt}
	\end{subfigure}%

	\begin{subfigure}[b]{0.5\textwidth}
		\centering
		\includegraphics[height=2.0in]{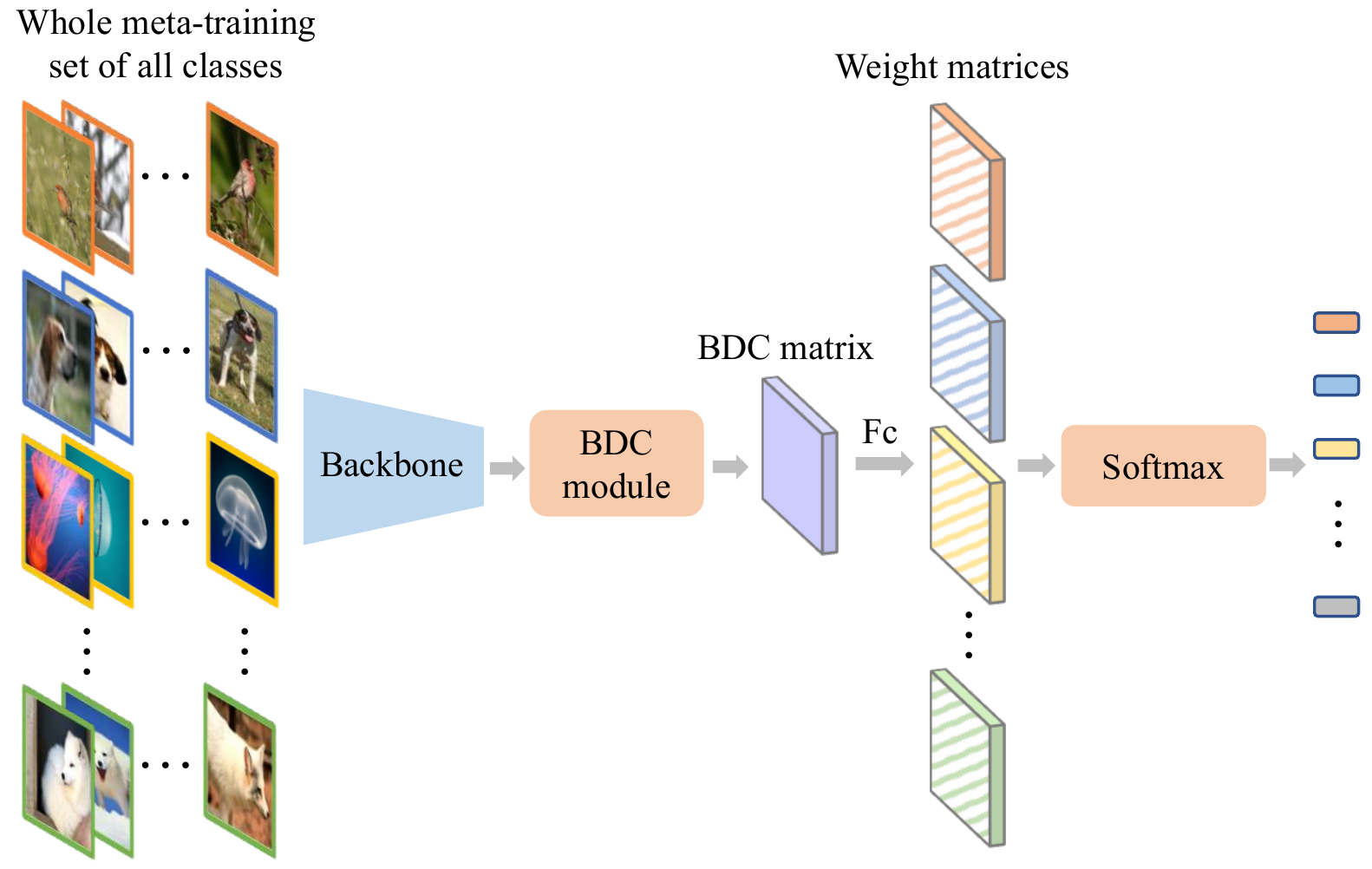}
		\caption{STL DeepBDC--Instantiation based on Good-Embed~\cite{tian2020rethinking} relying on non-episodic training. }
		\label{subfigure:overview_b}
	\end{subfigure}
	\caption{Two instantiations of our DeepBDC for few-shot classification.  Meta DeepBDC (a) is based on the idea of meta learning which depends  on  episodic training; here we take a 3-way 2-shot classification  as an example for illustration. In STL DeepBDC (b), we train a network with a conventional softmax classifier and cross-entroy loss  on the whole meta-training spanning all classes; during meta-testing, we use the trained network as an embedding model for feature extraction, constructing and training a logistic regression model for classification. }
	\label{fig:fusion schemes}
	\vspace{-8pt}
	\end{figure}

\subsection{Instantiating DeepBDC for Few-shot Learning}\label{subsection:instantiation}

We instantiate our DeepBDC   based on meta-learning framework and on simple transfer learning framework, and the resulting Meta DeepBDC and STL DeepBDC are shown in Fig.~\ref{subfigure:overveiw_a} and Fig.~\ref{subfigure:overview_b},  respectively.

\vspace{3pt}\noindent\textbf{Meta DeepBDC} Standard few-shot  learning is performed in an episodic manner  on a multitude of tasks. A  task is often formulated as  a $N$-way $K$-shot classification problem, which spans $N$ classes each with $K$ support images and $Q$ query images, on a support set $\mathcal{D}^{\mathrm{sup}}=\{(\mathbf{z}_{j},y_{j})\}_{j=1}^{NK}$ and a query set $\mathcal{D}^{\mathrm{que}}=\{(\mathbf{z}_{j},y_{j})\}_{j=1}^{NQ}$. A learner is trained on $\mathcal{D}^{\mathrm{sup}}$ and makes predictions on $\mathcal{D}^{\mathrm{que}}$. 

We instantiate Meta DeepBDC with ProtoNet~\cite{snell2017prototypical} as a blue print. It learns a metric space where classification is performed by computing distances to the prototype of every class.
On one task $(\mathcal{D}^{\mathrm{sup}},\mathcal{D}^{\mathrm{que}})$, we feed  image  $\mathbf{z}_{j}$ to the network to produce the BDC matrix $\mathbf{A}_{\boldsymbol{\uptheta}}(\mathbf{z}_{j})$. 
 The prototype of the support class $k$ is the average (Avg) of the BDC matrices belonging to its class:
\begin{align}
	\mathbf{P}_{k}=\frac{1}{K}\sum_{(\mathbf{z}_{j}, y_{j})\in \mathcal{S}_{k}}\mathbf{A}_{\boldsymbol{\uptheta}}(\mathbf{z}_{j})
\end{align} 
where $\mathcal{S}_{k}$ is the set of examples in $\mathcal{D}^{\mathrm{sup}}$ labeled with class $k$.
We produce a distribution over classes based on a softmax over distances to the prototypes of the support classes,  and then formulate the following loss function:
\begin{align}\label{equ:Meta DeepBDC loss}
\!\arg\min_{\boldsymbol{\uptheta}}\;\;-\!\!\!\!\!\!\sum_{(\mathbf{z}_{j}, y_{j})\in  \mathcal{D}^{\mathrm{que}}}\!\!\log\frac{\exp(\tau\mathrm{tr}(\mathbf{A}_{\boldsymbol{\uptheta}}(\mathbf{z}_{j})^{T}\mathbf{P}_{{y}_{j}}))}{\sum_{k}\exp(\tau\mathrm{tr}(\mathbf{A}_{\boldsymbol{\uptheta}}(\mathbf{z}_{j})^{T}\mathbf{P}_{k}))}
\end{align} 
where   $\tau$ is a learnable scaling parameter~\cite{Wertheimer_2021_CVPR,ye2020few,chen2021meta}.

We train the  learner by sampling  tasks   from  a massive meta-training set $\mathcal{C}^{\mathrm{train}}$ where the number of classes is far larger than $N$. Then, we sample tasks from a held-out meta-testing set $\mathcal{C}^{\mathrm{test}}$ on which we evaluate the performance of the  learner. The episodic training ensures consistency between meta-training and meta-testing, which is  crucial for the  meta-learning methods~\cite{vinyals2016matching,snell2017prototypical}.

\vspace{3pt}\noindent\textbf{STL DeepBDC} This instantiation is based on a widely used simple transfer learning (STL) framework~\cite{pmlr-v32-donahue14},  in which a deep network is trained on a large dataset and  is then used as an embedding model to extract features for  downstream tasks with  few labeled examples.  

We train a conventional image classification task on the whole meta-training set $\mathcal{C}^{\mathrm{train}}$ spanning all classes. 
The cross-entropy loss between prediction and ground-truth labels is used for training a learner from scratch:
\begin{align}\label{equ:STL DeepBDC loss}
\!\!\arg\min_{\boldsymbol{\uptheta},\mathbf{W}_{k}}\;\;-\!\!\!\!\!\!\!\!\sum_{(\mathbf{z}_{j}, y_{j})\in  \mathcal{C}^{\mathrm{train}}}\!\!\!\!\log\frac{\exp(\tau\mathrm{tr}(\mathbf{A}_{\boldsymbol{\uptheta}}(\mathbf{z}_{j})^{T}\mathbf{W}_{{y}_{j}}))}{\sum_{k}\exp(\tau\mathrm{tr}(\mathbf{A}_{\boldsymbol{\uptheta}}(\mathbf{z}_{j})^{T}\mathbf{W}_{k}))}
\end{align} 
where $\mathbf{W}_{k}\in \mathbb{R}^{d\times d}$ is the $k$-th weight matrix and $\tau$ is a learnable scaling parameter. 
 For a task $(\mathcal{D}^{\mathrm{sup}},\mathcal{D}^{\mathrm{que}})$ sampled from meta-testing set  $\mathcal{C}^{\mathrm{test}}$, we build and train a new linear classifier for $K$ classes on $\mathcal{D}^{\mathrm{sup}}$, using the trained model as a feature extractor. Following~\cite{tian2020rethinking}, we adopt the  logistic regression model for classification, and,  instead of directly using the trained model for meta-testing tasks,  a sequential self-distillation technique is used to distill  knowledge from the trained model on the meta-training set. 

By referring to Eq.~(\ref{equ:Meta DeepBDC loss}) and Eq.~(\ref{equ:STL DeepBDC loss}), we can interpret  $\mathbf{W}_{k}$ as the prototype of class $k$, a dummy BDC matrix learned through training. Note that similar interpretations are given in DeepEMD~\cite{zhang2020deepemd} and FRN~\cite{Wertheimer_2021_CVPR}.  It is worth mentioning that,  by vectorization operation as described in Eq.~(\ref{equ: BDC measure-vector}) for both the BDC matrices and the weight matrices, the softmax function in Eq.~(\ref{equ:STL DeepBDC loss}) can be implemented via a standard fully-connected (FC) layer. 

\subsection{Relation with  Previous Methods}\label{subsection: relation to previus methods}

Let $\{\mathbf{x}_{j}\}_{j=1}^{n}$ be  features  of a query image, viewed as the observations of a random vector $X$.  One can compute  the mean vector $\boldsymbol{\upmu}_{X}=\frac{1}{n}\sum_{j=1}^{n}\mathbf{x}_{j}$, covariance matrix $\boldsymbol{\Sigma}_{X}=\frac{1}{n}\sum_{j=1}^{n}(\mathbf{x}_{j}-\boldsymbol{\upmu}_{X})(\mathbf{x}_{j}-\boldsymbol{\upmu}_{X})^{T}$ or Gaussian distribution $\mathcal{N}_{\boldsymbol{\mu}_{X},\boldsymbol{\Sigma}_{X}}$, as image representations. Note that  these representations have  been extensively studied outside of the few-shot learning regime, where they are deemed global average pooling~\cite{He_2016_CVPR},  bilinear~\cite{Binear_PAMI2018} or covariance pooling~\cite{MPN-COV_PAMI2021}, and  Gaussian pooling~\cite{G2DeNet_CVPR2017}, respectively.  The corresponding  prototypes  of the support class, $\boldsymbol{\upmu}_{Y}$, $\boldsymbol{\Sigma}_{Y}$ or $\mathcal{N}_{\boldsymbol{\mu}_{Y},\boldsymbol{\Sigma}_{Y}}$,  can be computed using the features of $K$ support images.

\vspace{2pt}\noindent\textbf{ProtoNet}~\cite{snell2017prototypical}  represents the images with the mean vector and measures the difference using  Euclidean distance 
$\rho_{\mathrm{ProtoNet}}(X,Y)\!\!=\!\!\|\boldsymbol{\upmu}_{X}\!\!-\!\!\boldsymbol{\upmu}_{Y}\|^{2}$ or cosine similarity ${\boldsymbol{\upmu}_{X}^{T}\boldsymbol{\upmu}_{Y}}/(\|\boldsymbol{\upmu}_{X}\|\|\boldsymbol{\upmu}_{Y}\|)$ 
for metric learning.

\vspace{2pt}\noindent\textbf{CovNet}~\cite{Wertheimer2019}  adopts the covariance matrices as image representations for improving the first-order representation. The covariance matrices are  subject to signed square-root normalization and then are compared with the Euclidean distance in the matrix space (i.e., the Frobenius norm) 
$\rho_{\mathrm{CovNet}}(X,Y)=\|\boldsymbol{\Sigma}_{X}-\boldsymbol{\Sigma}_{Y}\|^{2}$.

\vspace{2pt}\noindent\textbf{ADM}~\cite{ADM}  proposes to use an asymmetric distribution measure (ADM) to evaluate the dis-similarity between the query image and the support class. The distributions of images are represented by multivariate Gaussians whose differences are measured by  KL-divergence $\rho_{\mathrm{ADM}}(X,Y)=D_{\mathrm{KL}}(\mathcal{N}_{\boldsymbol{\mu}_{X},\boldsymbol{\Sigma}_{X}}||\mathcal{N}_{\boldsymbol{\mu}_{Y},\boldsymbol{\Sigma}_{Y}})$.

\vspace{2pt}\noindent\textbf{DeepEMD}~\cite{zhang2020deepemd} uses discrete distributions as image representations. Specifically, the discrete PDF of the query image is  $f_{X}(\mathbf{x})\!\!=\!\!\sum_{j=1}^{n}\!f_{\mathbf{x}_{j}}\delta_{\mathbf{x},\mathbf{x}_{j}}$, where $f_{\mathbf{x}_{j}}$ denotes the probability of  $\mathbf{x}_{j}$ and $\delta_{\mathbf{x},\mathbf{x}_{j}}$ is the Kronecker delta  which is equal to  1 if $\mathbf{x}\!\!=\!\!\mathbf{x}_{j}$ and  zero otherwise. Let  the PDF of a support image be  $f_{Y}(\mathbf{y})\!\!=\!\!\sum_{j=1}^{n}\!f_{\mathbf{y}_{j}}\delta_{\mathbf{y},\mathbf{y}_{j}}$. The distance between $f_{X}(\mathbf{x})$ and $f_{Y}(\mathbf{y})$ is formulated as EMD,  i.e.,  $\rho_\mathrm{EMD}(X,Y)\!\!=\!\!\min_{f_{\mathbf{x}_j,\mathbf{y}_l}\geq 0}\sum_{j=1}^{n}\!\sum_{l=1}^{n}\!f_{\mathbf{x}_j,\mathbf{y}_l}c_{\mathbf{x}_j,\mathbf{y}_l}$ with  constraints  $\sum_{l=1}^{n}\!f_{\mathbf{x}_j,\mathbf{y}_l}\!\!=\!\!f_{\mathbf{x}_{j}}$ and $\sum_{j=1}^{n}\!f_{\mathbf{x}_j,\mathbf{y}_l}\!\!=\!\!f_{\mathbf{y}_{l}}$ for $j,l\!=\!1,\ldots, n$. Here $c_{\mathbf{x}_j,\mathbf{y}_l}$ is the transport cost.  Thus, EMD  seeks an optimal joint distribution $f_{XY}(\mathbf{x}_j,\mathbf{y}_l)\!\stackrel{\scriptscriptstyle\vartriangle}{=}\!f_{\mathbf{x}_j,\mathbf{y}_l}$ such  that the expectation of transportation cost  is minimal~\cite[Sec. 2.3]{MAL-073}. DeepEMD proposes a  cross-reference mechanism to define $f_{\mathbf{x}_{j}}$ and $f_{\mathbf{y}_{l}}$, and a structured FC layer to handle $K$-shot classification ($K$$>$$1$).

\begin{table*}[htb!]
	\centering
	\setlength{\tabcolsep}{3pt}
	\footnotesize
	\begin{tabular}{ll}
		\parbox{3.2in}{
			\begin{minipage}{1\linewidth}
				\begin{subtable}{1\linewidth}
					\centering
					\setlength\tabcolsep{3pt}
					\renewcommand{\baselinestretch}{1.0}
					\begin{tabular}{c|c|c|c|c|c}
						\hline
						\multirow{2}*{$d$}  & Parameters & \multicolumn{2}{c|}{1-shot} & \multicolumn{2}{c}{5-shot}   \\
						\cline{3-6}
						& (M) & Acc & Latency & Acc & Latency \\
						\hline
					   1280 & 13.25   &66.36$\pm$0.43 &488& 83.23$\pm$0.30  &614\\
						960 &  13.04  &66.81$\pm$0.44 &280& 83.68$\pm$0.28  &351\\
						640 & 12.84  &\textbf{67.34$\pm$0.43} &161& \textbf{84.46$\pm$0.28}  &198\\
						512 & 12.75  &67.10$\pm$0.45 &134& 84.23$\pm$0.28&164  \\
						256 & 12.59 &66.90$\pm$0.43 &  121&84.15$\pm$0.28  &148 \\
						\hline
						\multicolumn{2}{c|}{ProtoNet~\cite{snell2017prototypical}} & 62.11$\pm$0.44 &115& 80.77$\pm$0.30 & 143 \\
						\hline
						\hline
						\multicolumn{2}{c|}{\multirow{2}*{Similarity function}} & \multicolumn{2}{c|}{1-shot} & \multicolumn{2}{c}{5-shot} \\
						\cline{3-6}
						\multicolumn{2}{c|}{} & Acc & Latency & Acc & Latency \\
						\hline
						\multicolumn{2}{c|}{Inner product} & \textbf{67.34$\pm$0.43} & 161&82.38$\pm$0.32 &193\\
						\multicolumn{2}{c|}{Cosine similarity} & 61.74$\pm$0.42 &172 &82.49$\pm$0.31 &207\\
						\multicolumn{2}{c|}{Euclidean distance } &56.70$\pm$0.45 &163 &\textbf{84.46$\pm$0.28}  &198\\
						\hline
					\end{tabular}
					\caption{Meta DeepBDC based on ProtoNet~\cite{snell2017prototypical} as a blueprint.}
					\label{subtable: Meta DeepBDC}
				\end{subtable}%
			\end{minipage}
		}&
		\hspace{0em}
		\parbox{3.5in}{
			\begin{minipage}{1\linewidth}
				\begin{subtable}{1\linewidth}
					\centering
					\setlength\tabcolsep{3pt}
					\renewcommand{\baselinestretch}{1.0}
					\begin{tabular}{c|c|c|c|c|c}
						\hline
						\multirow{2}*{$d$}  & Parameters & \multicolumn{2}{c|}{1-shot} & \multicolumn{2}{c}{5-shot}   \\
						\cline{3-6}
						& (M) & Acc & Latency & Acc & Latency \\
						\hline
						512 & 13.41  &  64.92$\pm$0.43 &1110 & 84.61$\pm$0.29 &2016 \\
						256 & 12.75  &  66.15$\pm$0.43 &371& 85.44$\pm$0.29  &587  \\
						196 & 12.65  & 66.57$\pm$0.43 &285& 85.36$\pm$0.29  &424  \\
						128 & 12.55 &  \textbf{67.83$\pm$0.43} &184& \textbf{85.45$\pm$0.30} &245\\
						64 & 12.48  &66.97$\pm$0.44 &137 & 83.18$\pm$0.30 & 172\\
						\hline
						\multicolumn{2}{c|}{Good-Embed~\cite{tian2020rethinking}} & 64.82$\pm$0.44 &121& 82.14$\pm$0.43 & 155\\ 
						\hline
						\hline
						\multicolumn{2}{c|}{\multirow{2}*{Classifier}} & \multicolumn{2}{c|}{1-shot} & \multicolumn{2}{c}{5-shot}  \\
						\cline{3-6}
						\multicolumn{2}{c|}{} & Acc & Latency & Acc & Latency \\
						\hline
						\multicolumn{2}{c|}{Logistic regression} &\textbf{67.83$\pm$0.43} &184& \textbf{85.45$\pm$0.30} &245 \\
						\multicolumn{2}{c|}{SVM} & 66.29$\pm$0.44&113& 84.73$\pm$0.29&144\\
						\multicolumn{2}{c|}{Softmax} & 66.30$\pm$0.44&1250& 85.20$\pm$0.29&4374 \\
						\hline
					\end{tabular}
					\caption{STL DeepBDC based on~\cite{tian2020rethinking} relying on non-episodic training.}
					\label{subtable: STL DeepBDC}
				\end{subtable}%
			\end{minipage}
		}
	\end{tabular}
	\caption{Ablation analysis of  our two instantiations of DeepBDC with the backbone of ResNet-12 on \textit{mini}ImageNet. We report  accuracy  and latency (ms) of   one meta-testing task for 5-way classification. Latency is measured with a GeForce GTX 1080. }\label{table: Meta and STL DeepBDC}
\end{table*}

\section{Experiments}

We first describe briefly the experimental settings. Next, we perform  ablation study on our two instantiations (i.e., Meta DeepBDC and STL DeepBDC) and  make comparisons to the counterparts.  Finally, we compare  with  state-of-the-art methods on six few-shot  datasets, covering general object recognition, fine-grained   categorization and cross-domain  classification. 

\begin{table}
	\centering
	\footnotesize
	\setlength\tabcolsep{4pt}
	\renewcommand{\baselinestretch}{1.0}
	\begin{tabular}{c|c|c|c|c}
		\hline
		\multirow{2}*{Method}  &  \multicolumn{2}{c|}{1-shot} & \multicolumn{2}{c}{5-shot} \\
		\cline{2-5}
		& Acc & Latency & Acc & Latency \\
		\hline
		ProtoNet~\cite{snell2017prototypical} & 62.11$\pm$0.44 & \textbf{115} &  80.77$\pm$0.30  &\textbf{143} \\
		ADM~\cite{ADM} &65.87$\pm$0.43 &199 & 82.05$\pm$0.29 &221  \\
		CovNet~\cite{Wertheimer2019} &64.59$\pm$0.45 &120 &  82.02$\pm$0.29 &144  \\
		DeepEMD~\cite{zhang2020deepemd} & 65.91$\pm$0.82 & 457 &  82.41$\pm$0.56 &12617 \\
		\hline
		Meta DeepBDC & 67.34$\pm$0.43 & 161 & 84.46$\pm$0.28 & 198  \\ 
		STL DeepBDC & \textbf{67.83}$\pm$0.43 & 184 & \textbf{85.45$\pm$0.29} & 245  \\ 
		\hline
	\end{tabular}
	\caption{Comparison of accuracy and latency (ms)   of 5-way classification to the counterparts with the backbone of ResNet-12 on \textit{mini}ImageNet.}
	\label{tab:resolution} 	
\end{table}

\subsection{Experimental Settings}\label{sec:implementation}

\vspace{2pt}\noindent\textbf{Datasets} We  experiment on two general object recognition benchmarks, i.e.,  \textit{mini}ImageNet~\cite{vinyals2016matching} and \textit{tiered}ImageNet~\cite{ren2018meta},  and one fine-grained image classification dataset, i.e.,  CUB-200-2011~\cite{wah2011caltech} (CUB for short). We also evaluate domain transfer ability of models by training on \textit{mini}ImageNet and then test on CUB~\cite{wah2011caltech},  Aircraft~\cite{maji13fine-grained} and Cars~\cite{Krause20133D}. 

\vspace{2pt}\noindent\textbf{Backbone network} For fair comparisons with previous methods, we use two kinds of networks as backbones, i.e., ResNet-12~\cite{tian2020rethinking,MetaOptNet} and ResNet-18~\cite{afrasiyabi2020associative,liu2020negative,sung2018learning}. Same as commonly used practice, the input resolution of images is 84$\times$84 for ResNet-12 and 224$\times$224 for ResNet-18, respectively. Moreover, we  adopt deeper models with higher capacity, i.e.,  ResNet-34~\cite{He_2016_CVPR} with input images of 224$\times$224 and a variant of ResNet-34 fit for input images of 84$\times$84.   Similar to~\cite{CTX,ADM},  we remove the last down-sampling of  backbones to  obtain more convolutional features. 

\vspace{2pt}\noindent\textbf{Training}  Our Meta DeepBDC is based on meta-learning framework, depending on episodic training.  Each episode (task)  concerns standard 5-way 1-shot or 5-way 5-shot classification, uniformly sampled from meta-training or meta-testing set; following~\cite{zhang2020deepemd,Wertheimer_2021_CVPR,chen2021meta}, before episodic training, we pre-train the models whose weights are used as  initialization. Contrary  to Meta DeepBDC, our STL DeepBDC is based on simple transfer learning framework, requiring non-episodic training. Following~\cite{tian2020rethinking}, we train a network  as an embedding model with cross-entropy loss on the whole meta-training set spanning all classes; for each meta-testing task, we train a new logistic regression classifier using the  features extracted by the  embedding model.

In supplement (Supp.) S1, we provide statistics of  datasets and the splits of meta training/validation/test sets,  as well as   details  on network architectures, optimizers, hyperparameters, etc. 

\subsection{Ablation Study}

\begin{table*}[htb!]
	\centering
	\setlength{\tabcolsep}{3pt}
	\footnotesize
	\begin{tabular}{ll}
		\parbox{4.0in}{
			\begin{minipage}{1\linewidth}
				\begin{subtable}[t]{0.3\linewidth}
					\centering
					\footnotesize
					\setlength\tabcolsep{2.5pt}
					\renewcommand{\baselinestretch}{1.0}
					\captionsetup{width=3.1\textwidth}
					\renewcommand{\arraystretch}{1.0}
					\begin{tabular}{lrcccc}
						\hline
						\multirow{2}*{Method}   & \multirow{2}*{Backbone} & \multicolumn{2}{c}{\textbf{\textit{mini}ImageNet}} & \multicolumn{2}{c}{\textbf{\textit{tiered}ImageNet}}   \\
						&  & 1-shot & 5-shot & 1-shot & 5-shot   \\ \hline  
						CTM~\cite{li2019ctm} & ResNet-18 & 64.12$\pm$0.82 & 80.51$\pm$0.13 & 68.41$\pm$0.39 & 84.28$\pm$1.73 \\
						S2M2~\cite{mangla2020charting} & ResNet-18 & 64.06$\pm$0.18 & 80.58$\pm$0.12 & -- & -- \\
						TADAM~\cite{oreshkin2018tadam}   & ResNet-12 & 58.50$\pm$0.30 & 76.70$\pm$0.38 & -- & -- \\
						MetaOptNet~\cite{MetaOptNet}   & ResNet-12 &  62.64$\pm$0.44 & 78.63$\pm$0.46 & 65.99$\pm$0.72 & 81.56$\pm$0.63  \\
						DN4~\cite{DN4-CVPR2019}~$^{\dagger}$ & ResNet-12 & 64.73$\pm$0.44 & 79.85$\pm$0.31 & -- &--   \\ 
						Baseline++~\cite{chen2019closer}~$^{\dagger}$ & ResNet-12 & 60.56$\pm$0.45 & 77.40$\pm$0.34 & -- & --   \\
						Good-Embed\cite{tian2020rethinking}   & ResNet-12 &  64.82$\pm$0.60 & 82.14$\pm$0.43 &71.52$\pm$0.69 &86.03$\pm$0.58  \\
						
						FEAT ~\cite{ye2020few}   & ResNet-12  &  66.78$\pm$0.20 & 82.05$\pm$0.14 &70.80$\pm$0.23 &84.79$\pm$0.16  \\
						Meta-Baseline~\cite{chen2021meta} & ResNet-12   & 63.17$\pm$0.23 & 79.26$\pm$ 0.17 & 68.62$\pm$0.27 &83.29$\pm$0.18  \\
						MELR~\cite{MELR}   & ResNet-12  &  \color{secondbest}{67.40$\pm$0.43} & 83.40$\pm$0.28 & 72.14$\pm$0.51 & 87.01$\pm$0.35 \\
						FRN~\cite{Wertheimer_2021_CVPR}   & ResNet-12 &  66.45$\pm$0.19 & 82.83$\pm$0.13 & 71.16$\pm$0.22 & 86.01$\pm$0.15  \\
						IEPT~\cite{IEPT}     & ResNet-12 &  67.05$\pm$0.44 & 82.90$\pm$0.30 &72.24$\pm$0.50 &86.73$\pm$0.34 \\
						BML~\cite{zhou2021bml}  & ResNet-12  &  67.04$\pm$0.63 & 83.63$\pm$0.29 &68.99$\pm$0.50 &85.49$\pm$0.34  \\
						ProtoNet~\cite{snell2017prototypical}~$^{\dagger}$   &  ResNet-12 & 62.11$\pm$0.44 & 80.77$\pm$0.30 & 68.31$\pm$0.51 & 83.85$\pm$0.36 \\
						ADM~\cite{ADM}~$^{\dagger}$   & ResNet-12 & 65.87$\pm$0.43 & 82.05$\pm$0.29&70.78$\pm$0.52 &85.70$\pm$0.43 \\
						CovNet~\cite{Wertheimer2019}~$^{\dagger}$   & ResNet-12 & 64.59$\pm$0.45 & 82.02$\pm$0.29 & 69.75$\pm$0.52& 84.21$\pm$0.26 \\
						DeepEMD~\cite{zhang2020deepemd}  & ResNet-12 & 65.91$\pm$0.82 & 82.41$\pm$0.56 &71.16$\pm$0.87 & 86.03$\pm$0.58  \\
						\hline
						Meta DeepBDC  & ResNet-12 & 67.34$\pm$0.43 & \color{secondbest}{84.46$\pm$0.28}&\color{secondbest}{72.34$\pm$0.49} &\color{secondbest}{87.31$\pm$0.32}  \\
						STL DeepBDC  & ResNet-12 & 
						\textbf{67.83$\pm$0.43}&  \textbf{85.45$\pm$0.29}&  \textbf{73.82$\pm$0.47}& \textbf{89.00$\pm$0.30}  \\
						\hline
					\end{tabular}
					\caption{\label{tab:sota_general} Results on general object recognition datasets.}
				\end{subtable}
			\end{minipage}
		}&
		\hspace{0.1em}
		\parbox{3.5in}{
			\begin{minipage}{1\linewidth}
				\begin{subtable}[t]{0.3\linewidth}
					\footnotesize
					\setlength\tabcolsep{2.5pt}
					\captionsetup{width=2.2\textwidth}
					\begin{tabular}{lcll}
						\hline
						\multirow{2}*{Method} & \multirow{2}*{Backbone} & \multicolumn{2}{c}{\textbf{CUB}} \\
						& & 1-shot & 5-shot \\ 
						\hline
						ProtoNet~\cite{snell2017prototypical} & Conv4 & 64.42$\pm$0.48 & 81.82$\pm$0.35  \\
						FEAT~\cite{ye2020few} & Conv4 &  68.87$\pm$0.22 & 82.90$\pm$0.15      \\
						MELR~\cite{MELR} & Conv4  & 70.26$\pm$0.50 & 85.01$\pm$0.32 \\
						MVT~\cite{pmlr-v119-park20b} & ResNet-10 & \multicolumn{1}{c}{\centering --} & 85.35$\pm$0.55 \\
						MatchNet~\cite{vinyals2016matching} & ResNet-12 & 71.87$\pm$0.85 & 85.08$\pm$0.57 \\
						Wang \etal LR~\cite{WangICI2020} & ResNet-12 & 76.16  &  90.32   \\
						MAML~\cite{finn2017model}  & ResNet-18 &  68.42$\pm$1.07 & 83.47$\pm$0.62 \\
						$\Delta$-encoder~\cite{delta-encoder} & ResNet-18 & 69.80  & 82.60  \\
						Baseline++~\cite{chen2019closer}   & ResNet-18 & 67.02$\pm$0.90 & 83.58$\pm$0.54 \\
						AA~\cite{afrasiyabi2020associative}  & ResNet-18 & 74.22$\pm$1.09  &88.65$\pm$0.55 \\
						Neg-Cosine~\cite{liu2020negative}   & ResNet-18 &  72.66$\pm$0.85 & 89.40$\pm$0.43 \\
						LaplacianShot~\cite{Ziko2020}  & ResNet-18 &  80.96  & 88.68   \\
						FRN~\cite{Wertheimer_2021_CVPR}~$^{\dagger}$   & ResNet-18 & 82.55$\pm$0.19 & 92.98$\pm$0.10 \\
						Good-Embed\cite{tian2020rethinking}~$^{\dagger}$   & ResNet-18 & 77.92$\pm$0.46& 89.94$\pm$0.26\\
						ProtoNet~\cite{snell2017prototypical}~$^{\dagger}$   & ResNet-18 &80.90$\pm$0.43 & 89.81$\pm$0.23\\
						ADM~\cite{ADM}~$^{\dagger}$ & ResNet-18 & 79.31$\pm$0.43 & 90.69$\pm$0.21 \\
						CovNet~\cite{Wertheimer2019}~$^{\dagger}$ & ResNet-18 & 80.76$\pm$0.42 & 92.05$\pm$0.20 \\
						\hline
						Meta DeepBDC & ResNet-18 & \color{secondbest}{83.55$\pm$0.40} &  \color{secondbest}{93.82$\pm$0.17} \\
						STL DeepBDC & ResNet-18 &
						\textbf{84.01$\pm$0.42}  & \textbf{94.02$\pm$0.24} \\
						\hline
					\end{tabular}
					\caption{\label{tab:CUB}Results on fine-grained categorization dataset.}
				\end{subtable}
			\end{minipage}
		}
	\end{tabular}
	\caption{Comparison with state-of-the-art methods for both  general  and  fine-grained few-shot image classification. The best results are in \textbf{bold black} and second-best ones are in {\color{secondbest}{red}}. $^{\dagger}$ Reproduced  with our setting. }\label{table:sota}
	\vspace{-8pt}
\end{table*}

We perform  ablation analysis of our two instantiations and compare to the counterparts  on \textit{mini}ImageNet for 5-way task, with  ResNet-12 as the backbone.  Additional details on  implementation of the counterparts and  extra experiments are respectively given in Supp. S-2 and Supp. S-3.

\vspace{2pt}\noindent \textbf{Ablation analysis of Meta DeepBDC}  
As the sizes of BDC matrices are quadratic in the number of channels, we introduce a $1$$\times$$1$ convolution (conv) layer, decreasing   the channel number to $d$. In our implementation,  each BDC matrix is vectorized as in Eq.~(\ref{equ: BDC measure-vector}), thus is of size $d(d+1)/2$.  Tab.~\ref{subtable: Meta DeepBDC} (top) shows the effect of varying  $d$ on accuracy and  on  meta-testing time per episode.   We can see that the highest accuracy is achieved  when  $d\!=\!640$; meanwhile, the meta-testing time only  increases  moderately as $d$ enlarges. We also experiment by directly attaching BDC module to the backbone without the additional $1\!\times\!1$ conv layer; we achieve  67.10$\pm$0.43 and 84.50$\pm$0.28 for 1-shot and 5-shot, respectively, comparable to the best result  obtained by using  the additional  $1\!\times\!1$ conv layer.     Besides the inner product as depicted in Eq.~(\ref{equ: BDC measure-vector}), we can also use Euclidean distance or  cosine similarity as metric, and  the corresponding results are given in  Tab.~\ref{subtable: Meta DeepBDC} (bottom).  It can be seen that the inner product performs best for 1-shot task, while the Euclidean distance achieves the highest accuracy for 5-shot. The optimal setting achieved here is used  throughout  the remaining paper.  Finally, we note that Meta DeepBDC has much better performance than the  baseline (i.e., ProtoNet), regardless of the value of $d$, while increase of latency is small.

\vspace{2pt}\noindent \textbf{Ablation analysis of STL DeepBDC}  For each meta-testing task of STL DeepBDC, we need to build and train a new linear classifier which introduces parameters and computations.  As the size of BDC matrix is quadratic in $d$, the number of parameters  is considerable relative to that of training examples, particularly for larger $d$. Therefore, with increase of $d$, there exist greater risk of overfitting, which may explain why overall the accuracy becomes lower when  $d$ is larger  for both  1-shot and 5-shot tasks, as shown in Tab.~\ref{subtable: STL DeepBDC} (top). STL DeepBDC with $d\!=\!128$  obtains the best result,  higher than the best result of  Meta DeepBDC while taking comparable time.  Besides the logistic regression model, we  compare with softmax classifier and linear SVM as well.  From the results in Tab.~\ref{subtable: STL DeepBDC} (bottom), we can see that the softmax classifier is on par with SVM, while both of them are inferior to the logistic regression; the latency of logistic regression is larger than SVM while  the softmax classifier takes remarkably larger time than the other two.   At last,  we mention that STL DeepBDC with $d\!=\!128$ outperforms the baseline of Good-Embed by a large margin with moderate increase of latency. 

\vspace{2pt}\noindent \textbf{The i.i.d. assumption underlying DeepBDC}\label{i.i.d.}  The BDC metric depends on the i.i.d. assumption~\cite{Szekely2009} which is common in statistics and machine learning. As  in Sec.~\ref{subsection:BDC-computation}, by viewing each channel  (feature map) as a random observation,   we obtain  $d\!\times\!d$ matrices corresponding to the spatial pooling. Alternatively, one can regard  each spatial feature as a random observation,  leading to  $hw\!\times\!hw$ matrices which correspond to a channel pooling; however, for 1-shot/5-shot task and with the same setting as in  Tab.~\ref{subtable: Meta DeepBDC}, this   produces $62.55$$\pm$$0.45$/$78.88$$\pm$$0.32$ and $63.95$$\pm$$0.45$/$79.45$$\pm$$0.32$ for Meta DeepBDC ($d\!=\!640$) and STL DeepBDC ($d\!=\!128$), respectively, much lower than the  accuracies of spatial pooling. Note that the i.i.d. assumption may not hold for either spatial or channel pooling; our comparison suggests  the spatial pooling  is a better option. 

\vspace{2pt}\noindent \textbf{Comparison to the counterparts}
Here we compare with the counterparts whose representations are based on distribution modeling. Like our DeepBDC,  both ADM and CovNet  need to estimate second moments, which leads  to quadratic increase of representations. Therefore,  for a fair comparison with them,  we also add a $1$$\times$$1$ convolution with $d$ channels for dimension reduction, obtaining the best results for them by tuning $d$. The comparion results are presented in Tab.~\ref{tab:resolution}.

\begin{table*}[htb!]
	\centering
	\setlength{\tabcolsep}{3pt}
	\footnotesize
	\begin{tabular}{lll}
		\parbox{2.2in}{
			\begin{minipage}{1\linewidth}
				\begin{subtable}[t]{1\linewidth}
					\centering
					\footnotesize
					\setlength\tabcolsep{3pt}
					\begin{tabular}{ccc}
						\hline
						Method & Backbone   & 5-shot \\
						\hline
						Baseline~\cite{chen2019closer}  & ResNet-18  & 65.57$\pm$0.70 \\
						Baseline++~\cite{chen2019closer}  & ResNet-18  & 62.04$\pm$0.76 \\
						GNN+FT~\cite{tseng2020cross}  & ResNet-12  & 66.98$\pm$0.68 \\
						BML~\cite{zhou2021bml}  & ResNet-12  & 72.42$\pm$0.54 \\
						FRN~\cite{Wertheimer_2021_CVPR}  & ResNet-12  & 77.09$\pm$0.15 \\
						ProtoNet~\cite{snell2017prototypical}~$^{\dagger}$    & ResNet-12  & 67.19$\pm$0.38\\
						Good-Embed~\cite{tian2020rethinking}~$^{\dagger}$    & ResNet-12  & 67.43$\pm$0.44\\
						ADM~\cite{ADM}~$^{\dagger}$ & ResNet-12  &70.55$\pm$0.43 \\
						CovNet~\cite{Wertheimer2019}~$^{\dagger}$   & ResNet-12 & 76.77$\pm$0.34 \\
						\hline
						Meta DeepBDC & ResNet-12 &  \color{secondbest}{77.87$\pm$0.33}  \\
						STL DeepBDC & ResNet-12 &  \textbf{80.16$\pm$0.38} \\
						\hline
						
					\end{tabular}
					\caption{\label{tab:generalization_cub}\textit{mini}ImageNet $\rightarrow$ CUB.}
				\end{subtable}
			\end{minipage}
		}&
		\hspace{0em}
		\parbox{2.2in}{
			\begin{minipage}{1\linewidth}
				\begin{subtable}[t]{1\linewidth}
					\centering
					\footnotesize
					\setlength\tabcolsep{3pt}
					\begin{tabular}{ccc}
						\hline
						Method  & Backbone & 5-shot \\
						\hline
						ProtoNet~\cite{snell2017prototypical}~$^{\dagger}$   & ResNet-12  & 55.96$\pm$0.38   \\
						ADM~\cite{ADM}~$^{\dagger}$   & ResNet-12  &  65.40$\pm$0.36  \\
						CovNet~\cite{Wertheimer2019}~$^{\dagger}$   & ResNet-12 &   63.56$\pm$0.37\\
						Baseline~\cite{chen2019closer}~$^{\dagger}$    & ResNet-12  &   59.04$\pm$0.36  \\
						Baseline++~\cite{chen2019closer}~$^{\dagger}$    & ResNet-12  &   56.50$\pm$0.38  \\
						Good-Embed~\cite{tian2020rethinking}~$^{\dagger}$ & ResNet-12 & 58.95$\pm$0.38\\
						\hline
						Meta DeepBDC & ResNet-12 &  \color{secondbest}{68.67$\pm$0.39}   \\
						STL DeepBDC & ResNet-12 &  \textbf{69.07$\pm$0.39}   \\
						\hline
						
					\end{tabular}
					\caption{\label{tab:generalization_aircraft} \textit{mini}ImageNet $\rightarrow$ Aircraft.}
				\end{subtable}
			\end{minipage}
		} &
		\parbox{2.2in}{
			\begin{minipage}{1\linewidth}
				\begin{subtable}[t]{1\linewidth}
					\centering
					\footnotesize
					\setlength\tabcolsep{3pt}
					\begin{tabular}{ccc}
						\hline
						Method  & Backbone & 5-shot \\
						\hline
						ProtoNet~\cite{snell2017prototypical}~$^{\dagger}$   & ResNet-12  &  46.30$\pm$0.36  \\
						ADM~\cite{ADM}~$^{\dagger}$   & ResNet-12  &  53.94$\pm$0.35  \\
						CovNet~\cite{Wertheimer2019}~$^{\dagger}$   & ResNet-12 &   52.90$\pm$0.37 \\
						Baseline~\cite{chen2019closer}~$^{\dagger}$    & ResNet-12  &  50.29$\pm$0.37   \\
						Baseline++~\cite{chen2019closer}~$^{\dagger}$    & ResNet-12  &  46.44$\pm$0.37   \\
						Good-Embed~\cite{tian2020rethinking}~$^{\dagger}$ & ResNet-12 &  50.18$\pm$0.37 \\
						\hline
						Meta DeepBDC & ResNet-12 &  \color{secondbest}{54.61$\pm$0.37}  \\
						STL DeepBDC & ResNet-12 & \textbf{58.09$\pm$0.36}   \\
						\hline
						
					\end{tabular}
					\caption{\label{tab:generalization_cars}\textit{mini}ImageNet $\rightarrow$ Cars.}
				\end{subtable}
			\end{minipage}
		}
	\end{tabular}
	\caption{Comparison with state-of-the-art methods for 5-way 5-shot classification in cross-domain scenarios. The best results are in \textbf{bold black} and second-best ones are in {\color{secondbest}{red}}. $^{\dagger}$ Reproduced  with our  setting.  }\label{table:generalization}
	\vspace{-8pt}
\end{table*}

Regarding the \textit{accuracy},  we have several observations.   (1) ProtoNet  is inferior to CovNet and ADM, suggesting that second moments have better capability to model marginal distributions than  first moment.  (2) DeepEMD outperforms CovNet and ADM, which indicates that joint distribution modeling via EMD is superior to  modeling of marginal distributions.   (3)  Both our two instantiations outperform the counterparts by large margins. We attribute this to  that BDC has  stronger capability of statistical dependency modeling by effectively harnessing joint distributions. 
As to the \textit{latency},  our two instantiations both take a little longer time than ProtoNet and CovNet, while being comparable to ADM. Notably, DeepEMD is computationally  expensive, $\sim$ 2 times and  50 times slower than the other methods for 1-shot and 5-shot tasks, respectively.

\subsection{Comparison with State-of-the-art Methods}

\vspace{2pt}\noindent \textbf{General object recognition}  According to  Tab.~\ref{tab:sota_general}, on \textit{mini}ImageNet, for 1-shot task   Meta DeepBDC is on par with state-of-the-art MELR  while STL DeepBDC is  better than it; for 5-shot task,  Meta DeepBDC and STL DeepBDC outperform BML, which previously achieved the best result, by 0.83 percentage points (abbreviated as pp hereafter) and 1.82 pp, respectively. Our Meta DeepBDC can be further improved by combining  Image-to-Class Measure (DN4) following the idea introduced in~\cite{ADM}; accordingly, we achieve  67.86$\pm$0.41/85.14$\pm$0.29 for 1-shot/5-shot tasks, outperforming  ADM+DN4 (66.53$\pm$0.43/82.61$\pm$0.30). 
On \textit{tiered}ImageNet, for 1-shot task  Meta DeepBDC is slightly better than state-of-the-art IEPT while  STL DeepBDC outperforms it  by 1.58 pp; for 5-shot task,  Meta DeepBDC achieves slight gains (0.3 pp) over state-of-the-art  MELR while  STL DeepBDC has much larger gains ($\sim$2.0 pp).

\vspace{2pt}\noindent \textbf{Fine-grained categorization} Following~\cite{chen2019closer,liu2020negative,afrasiyabi2020associative}, we conduct experiments on CUB with the original  raw images. We reproduce ProtoNet and Good-Embed which are our baselines as well as  FRN. From Tab.~\ref{tab:CUB}, we can see that reproduced ProtoNet and Good-Embed are competitive with previous  published accuracies, indicating our re-implemented baselines provide fair competition; moreover, our methods are high-ranking across the board, compared to state-of-the-art FRN, the gains of  Meta DeepBDC and STL DeepBDC are 1.0/1.5 pp and 0.8/1.0 pp for 1-shot/5-shot task, respectively. Furthermore, by adopting  ResNet-34 with 224$\times$224 input images, our methods further improve. Specifically, Meta DeepBDC and STL DeepBDC  achieve 85.25{{$\pm$}0.39}/94.31$\pm$0.17 and 84.69$\pm$0.43/94.33$\pm$0.21, respectively,  outperforming corresponding baselines of ProtoNet  (80.58$\pm$/90.11$\pm$0.26) and Good-Embed  (79.33$\pm$0.48/90.10$\pm$0.28).

\vspace{2pt}\noindent \textbf{Cross-domain classification}
Finally, we evaluate 5-way 5-shot classification  in   cross-domain scenarios, by training  on \textit{mini}ImageNet and testing on  three widely used fine-grained datasets. Except DeepEMD which is computationally prohibitive for us, we implement all  counterparts based on distribution modeling,  and Good-Embed on the three datasets, as well as Baseline and Baseline++~\cite{chen2019closer} on Aircraft and Cars. The results are shown in Tab.~\ref{table:generalization}.    On \textit{mini}ImageNet$\rightarrow$CUB,   CovNet is very competitive, only slightly inferior to FRN, and both of them  are much better than the other methods except ours.  Meta DeepBDC and STL DeepBDC outperform  the high-performing FRN by  0.8 pp and 3.1 pp, respectively.  On \textit{mini}ImageNet$\rightarrow$Aircraft, our two instantiations improve over all the other compared methods by more than 3.2 pp.  On \textit{mini}ImageNet$\rightarrow$Cars, ADM is superior among our competitors; compared to it,  Meta DeepBDC and STL DeepBDC achieve 0.7 pp and 4.2 pp higher accuracies, respectively. These comparisons demonstrate that our models have stronger domain transfer capability.

\section{Conclusion} 

In this paper, we propose  a deep Brownian Distance Covariance (DeepBDC) method for few-shot classification. DeepBDC can effectively learn  image representations by measuring, for the query and support images,  the discrepancy between the joint distribution of their embedded features and  product of the marginals.  The core of DeepBDC is formulated as a modular and efficient layer, which can be flexibly inserted  into deep networks,  suitable not only for meta-learning framework based on episodic training, but also for the simple transfer learning framework that relies on  non-episodic training. Extensive experiments have shown that our DeepBDC method performs much better than the counterparts, and furthermore,  sets  new state-of-the-art results on multiple general, fine-grained and cross-domain few-shot classification tasks.  Our work shows great  potential of BDC, a fundamental but overlooked technique,  and encourages its future applications in deep learning.

{\small
\bibliographystyle{ieee_fullname}
\bibliography{egbib}
}

\clearpage

\begin{appendices}
	\noindent{\Large\bf Supplementary Material}
	\vspace{4mm}
\renewcommand{\theequation}{S-\arabic{equation}}
\renewcommand\thetable{S-\arabic{table}}
\renewcommand\thefigure{S-\arabic{figure}}




\appendix
\setcounter{section}{0}
\setcounter{table}{0}
\setcounter{figure}{0}
\setcounter{equation}{0}
\setcounter{footnote}{0}
\renewcommand{\appendixname}{}
\maketitle

In this supplement, we  present  details on the implementations of our DeepBDC and the counterparts. Besides, we conduct experiments providing  additional  ablation study and  comparison.  
We  finally show BDC's ability to characterize non-linear dependence.

\section*{S1~ Implementations}

\subsection*{S1-1~ Benchmarks}

\vspace{4pt}\noindent\textbf{\textit{mini}ImageNet} The \textit{mini}ImageNet~[\hyperref[reference:vinyals2016matching-supp]{\textcolor{green}{S-22}}] is a few-shot benchmark constructed from ImageNet~[\hyperref[reference:deng2009imagenet-supp]{\textcolor{green}{S-6}}] for general object recognition. It consists of  100 classes each of which contains 600 images.  Following previous works~[\hyperref[reference:chen2019closer-supp]{\textcolor{green}{S-4}},~\hyperref[reference:Wertheimer_2021_CVPR-supp]{\textcolor{green}{S-25}},~\hyperref[reference:ye2020few-supp]{\textcolor{green}{S-26}}], we use the splits provided by~[\hyperref[reference:Sachin2017-supp]{\textcolor{green}{S-18}}], which involves 64 classes for meta-training, 16 classes for meta-validation and the remaining 20 classes for meta-testing.

\vspace{4pt}\noindent\textbf{\textit{tiered}ImageNet}   The \textit{tiered}ImageNet~[\hyperref[reference:ren2018meta-supp]{\textcolor{green}{S-19}}] is also a few-shot benchmark for general object recognition. It is  constructed from  ImageNet~[\hyperref[reference:deng2009imagenet-supp]{\textcolor{green}{S-6}}] as well, which, different from \textit{mini}ImageNet, considers  the hierarchical structure of ImageNet.  This dataset contains 608 classes from 34 super-classes  and  a total of  779,165 images. Among these classes,  20 super-classes (351 classes) are used for meta-training,  6  super-classes (97 classes) for meta-validation and the remaining  8  super-classes (160 classes) for meta-testing.

\vspace{4pt}\noindent\textbf{CUB} \textbf{C}altech-\textbf{U}CSD \textbf{B}irds-200-2011~[\hyperref[reference:wah2011caltech-supp]{\textcolor{green}{S-23}}] (CUB) dataset is a widely used fine-grained categorization benchmark. This dataset contains 200 bird classes with 11,788 images in total. We use the splits of ~[\hyperref[reference:chen2019closer-supp]{\textcolor{green}{S-4}}], in which the total classes are divided into 100/50/50 for meta-training/validation/testing.  Following~[\hyperref[reference:chen2019closer-supp]{\textcolor{green}{S-4}},~\hyperref[reference:liu2020negative-supp]{\textcolor{green}{S-14}},~\hyperref[reference:afrasiyabi2020associative-supp]{\textcolor{green}{S-1}}], we conduct experiments on CUB with the original  raw images,  instead of cropped images via annotated bounding boxes~[\hyperref[reference:ye2020few-supp]{\textcolor{green}{S-26}},~\hyperref[reference:zhang2020deepemd-supp]{\textcolor{green}{S-27}}]. 

\vspace{4pt} \noindent\textbf{\textit{mini}ImageNet $\rightarrow$ CUB} Chen \etal~[\hyperref[reference:chen2019closer-supp]{\textcolor{green}{S-4}}] build the cross-domain task for assessing the domain transfer ability of the models. In this setting, all 100 classes of \textit{mini}ImageNet are  used for  meta-training, while the models are evaluated on the  meta-testing set (50 classes) of CUB.

\vspace{4pt}\noindent\textbf{\textit{mini}ImageNet $\rightarrow$ Aircraft} Aircraft~[\hyperref[reference:maji13fine-grained-supp]{\textcolor{green}{S-15}}] contains 10,000 images from 100 classes. We  perform meta-training on the whole \textit{mini}ImageNet; we adopt the splits on Aircraft proposed by~[\hyperref[reference:Wertheimer_2021_CVPR-supp]{\textcolor{green}{S-25}}], where 25 classes are used for  meta-validation and 25 classes are  for  meta-testing. Same as~[\hyperref[reference:Wertheimer_2021_CVPR-supp]{\textcolor{green}{S-25}}], we conduct experiments with the   images cropped by using the  bounding box annotations.

\vspace{6pt}\noindent\textbf{\textit{mini}ImageNet $\rightarrow$ Cars} Stanford Cars~[\hyperref[reference:Krause20133D-supp]{\textcolor{green}{S-10}}] (Cars)  contains 196 classes and  a total 16,185 images. We follow the splits of~[\hyperref[reference:DN4-supp]{\textcolor{green}{S-13}}] to build the meta-validation set (17 classes) and meta-testing set (49 classes). Similar to the other cross-domain benchmarks, the full \textit{mini}ImageNet is used as meta-training set.

\begin{table*}
	\centering
	\footnotesize
	\setlength\tabcolsep{8pt}
	\renewcommand\arraystretch{1.3}
	\begin{tabular}{c|c|c|c|c|c}
		\hline
		Method   & Hyper-params &  \textit{mini}ImageNet & \textit{tiered}ImageNet & CUB & \textit{mini}ImageNet $\rightarrow$ CUB$/$Aircraft$/$Cars  \\
		\hline
		\multirow{3}*{Meta DeepBDC}   &    Initial LR & 1e-4 & 5e-5  & 1e-3  & 5e-5 \\
		& LR decay  & [40, 80]*0.1 & [70]*0.1 & [40, 80]*0.1 & [70]*0.1 \\
		&    Epochs & 100 & 100 & 100 & 100 \\
		\hline
		\multirow{3}*{STL DeepBDC} &   Initial LR & 5e-2 & 5e-2 & 5e-2 & 5e-2\\
		& LR decay  & [100,150]*0.1 & [40,70]*0.1 & [120,170]*0.1 & [100,150]*0.1 \\
		&  Epochs & 180 & 100 & 220 & 180 \\
		\hline
		
	\end{tabular}
	\caption{\label{tab:hyper_params} Hyperparameter settings of our DeepBDC.}
\end{table*}

\subsection*{S1-2~ Architectures}\label{subsection: architectures}
For fair comparisons with previous methods, we use two kinds of networks as backbones, i.e., ResNet-12 and ResNet-18. Following previous practice, the input resolution of images is $84\!\times \!84$ for ResNet-12, and $224\!\times\! 224$  for ResNet-18, respectively. Besides, we use higher capacity models, i.e., ResNet-34 with input images of  $224\!\times\! 224$ and its variant fit for input images of 84$\times$84.

As in~[\hyperref[reference:tian2020rethinking-supp]{\textcolor{green}{S-21}},~\hyperref[reference:MetaOptNet-supp]{\textcolor{green}{S-11}},~\hyperref[reference:oreshkin2018tadam-supp]{\textcolor{green}{S-16}},~\hyperref[reference:Wertheimer_2021_CVPR-supp]{\textcolor{green}{S-25}}], ResNet-12 consists of four stages each of which contains one residual  block. The widths of the residual blocks of the four stages  are [64, 160, 320, 640]. Each residual block has three 3$\times$3 convolution layers with a batch normalization and a 0.1 leaky ReLU.  Right after every block except the last one, a 2$\times$2 max pooling layer is used to  down-sample the feature maps. 
We adopt ResNet-18 and ResNet-34 proposed  in~[\hyperref[reference:He_2016_CVPR-supp]{\textcolor{green}{S-8}}] with the last down-sampling being removed.   
The architecture of the variant of ResNet-34 is same as that of ResNet-12, except that the numbers of residual blocks of the four stages are [2, 3, 4, 2], instead of [1, 1, 1, 1] in ResNet-12.

\subsection*{S1-3~ Training and Evaluation Protocols} 

\paragraph{Training} During training, following~[\hyperref[reference:chen2019closer-supp]{\textcolor{green}{S-4}},~\hyperref[reference:Wertheimer_2021_CVPR-supp]{\textcolor{green}{S-25}},~\hyperref[reference:zhang2020deepemd-supp]{\textcolor{green}{S-27}}], we use   standard data augmentation including random resized crop, color jittering  and random horizontal flip on all benchmarks. We adopt the  SGD algorithm with a  momentum of 0.9 and a weight decay of  5e-4 to train our proposed DeepBDC networks.  For ResNet-12, we  apply  DropBlock~[\hyperref[reference:DropBlock-supp]{\textcolor{green}{S-7}}] regularization during training as in~[\hyperref[reference:ye2020few-supp]{\textcolor{green}{S-26}},~\hyperref[reference:MetaOptNet-supp]{\textcolor{green}{S-11}},~\hyperref[reference:chen2021meta-supp]{\textcolor{green}{S-5}}]. We  tune the number of epochs and the scheduling of learning rate (LR) on  different benchmarks. 

For our \textit{Meta DeepBDC} which is based on meta-learning framework, we  train the model by uniformly sampling episodes (tasks) from the meta-training set.  Following previous works~[\hyperref[reference:chen2021meta-supp]{\textcolor{green}{S-5}},~\hyperref[reference:Wertheimer_2021_CVPR-supp]{\textcolor{green}{S-25}},~\hyperref[reference:zhang2020deepemd-supp]{\textcolor{green}{S-27}}], right before performing  the episodic training, we pre-train the networks on the full meta-training set whose weights are used as initialization.  To sample a  5-way 1-shot task, we randomly pick up 5 classes each with  1 support image and 16 query images selected at random.  Similarly, every 5-way 5-shot task contains 5 support images and 16 query images.  Tab.~\ref{tab:hyper_params} (upper part)  shows the hyper-parameter settings for training Meta DeepBDC on all benchmarks. Let us take  \textit{mini}ImageNet as an example:  the initial learning rate (LR) is 1e-4, which is divided by 10 at epoch 40 and 80, respectively, and training proceeds until epoch 100. We adopt the model achieving highest accuracy on the meta-validation for evaluation on the meta-testing set.

Our \textit{STL DeepBDC} is based on Good-Embed~[\hyperref[reference:tian2020rethinking-supp]{\textcolor{green}{S-21}}], which falls into simple transfer learning framework, not  requiring episodic  training.  In practice, we train the network for a common multi-way classification task,   using a soft-max classifier via a  standard cross-entropy loss on the whole meta-training set spanning all classes.  We set the batch size to  64 across all benchmarks. Furthermore, following~[\hyperref[reference:tian2020rethinking-supp]{\textcolor{green}{S-21}}], we conduct  sequential self-distillation to distill knowledge from the trained model. The networks thus obtained are used as embedding models for extracting features (i.e., outputs of the last convolution  layer of one network). The hyper-parameter settings for STL DeepBDC are shown in Tab.~\ref{tab:hyper_params} (bottom part).

\paragraph{Evaluation}  We  uniformly sample  episodes (tasks) from the meta-testing set to evaluate the models' performance. Following ~[\hyperref[reference:chen2019closer-supp]{\textcolor{green}{S-4}}], we  build 5-way 1-shot or 5-shot setting, respectively, both with  15 query images. We report  mean accuracy of 2,000 episodes with 95\% confidence intervals. Our Meta DeepBDC does not require additional training, so we directly performing testing.  Our STL DeepBDC needs to train a linear classifier for each episode using the trained model  for feature extraction.  We implement the logistic regression via L-BFGS-B algorithm and linear SVM via LIBSVM  based on  scikit-learn software package~[\hyperref[reference:scikit-learn-supp]{\textcolor{green}{S-17}}]. We perform  L2 normalization for the  features in logistic regression and SVM; furthermore, we  standardize the normalized features before fed to SVM. Implementation of the soft-max classifier is based on PyTorch, where we adopt SGD algorithm with a batch size of 4, a  momentum of 0.9, a weight decay of 1e-3 and a learning rate of 1e-2.

\section*{S2~ Implementation of  the Counterparts}

\vspace{3pt}\noindent\textbf{ProtoNet } We use  implementation of  Chen \etal~[\hyperref[reference:chen2019closer-supp]{\textcolor{green}{S-4}}] which is public available~\footnote{\href{https://github.com/wyharveychen/CloserLookFewShot/}{https://github.com/wyharveychen/CloserLookFewShot}}.

\vspace{3pt}\noindent\textbf{ADM}  We use the code released by the authors~[\hyperref[reference:ADM-supp]{\textcolor{green}{S-12}}]~\footnote{\href{https://github.com/WenbinLee/ADM}{https://github.com/WenbinLee/ADM}}. Differently, we add one 1$\times$1 convolution for dimension reduction before computing mean vectors and covariance matrices. 
As the original method   performs unsatisfactorily, we use a different one fusing ADM and DN4~[\hyperref[reference:DN4-supp]{\textcolor{green}{S-13}}]. Specifically, we  attach independently  the ADM branch and DN4 branch to the backbone, both with individually learnable scaling parameters. Next, the negative KL-divergence score   and Image-to-Class (I2C) score are separately fed to softmax and are then added for the final cross  entropy loss. We combine Meta DeepBDC and DN4 similarly. 

\vspace{3pt}\noindent\textbf{CovNet}  We mainly follow  implementation of the authors ~[\hyperref[reference:Wertheimer2019-supp]{\textcolor{green}{S-24}}]~\footnote{\href{https://github.com/daviswer/fewshotlocal}{https://github.com/daviswer/fewshotlocal}}. Practically, we have two differences:  (1) we introduce a 1$\times$1 convolution for dimension reduction, and (2) for 5-way 1-shot classification, we use the inner product as the metric, rather than  the Frobenious norm  which produces poor results.

For all of the aforementioned methods, we  remove the last down-sampling of the backbones.  Following the previous practice ~[\hyperref[reference:Wertheimer_2021_CVPR-supp]{\textcolor{green}{S-25}},~\hyperref[reference:chen2021meta-supp]{\textcolor{green}{S-5}},~\hyperref[reference:zhou2021bml-supp]{\textcolor{green}{S-28}}], we employ the weights of pre-trained models as initialization before performing episodic training.

\section*{S3~ Additional Experiments}

This section introduces additional experiments to  ablate our DeepBDC and to compare with the counterparts.

\subsection*{S3-1~ On prototype in Meta DeepBDC} \label{subsection: effect of higher capacity}

In Meta DeepBDC, for the 5-shot setting,  the prototype of a support class is computed as the average of BDC matrices of 5 support images belonging to this class.  Here, we evaluate two other options for computing the prototype.  (1) We  average features of 5 support images, and then the averaged features are  used to compute the BDC matrix as the prototype. (2)  We concatenate features of 5 support images for computing the BDC matrix as the prototype. These two methods achieve  accuracies (\%) of  82.36 and 83.74  on \textit{mini}ImageNet, respectively, which are  lower than the accuracy obtained by averaging 5  support BDC matrices (84.46). 

\begin{table}
	\centering
	\footnotesize
	\setlength\tabcolsep{1.2pt}
	\renewcommand\arraystretch{1.3}
	\begin{tabular}{l|c|c|c|c|c|c|c|c}
		\hline
		\multirow{2}*{Method}   & \multicolumn{2}{c|}{ProtoNet$^{\dagger}$}  &  \multicolumn{2}{c|}{Good-Embed$^{\dagger}$} &  \multicolumn{2}{c|}{Meta DeepBDC} &  \multicolumn{2}{c}{STL DeepBDC} \\
		\cline{2-9}
		& 1-shot & 5-shot & 1-shot & 5-shot & 1-shot & 5-shot & 1-shot & 5-shot  \\
		\hline 
		ResNet-12 & 62.11 & 80.77 & 64.98  & 82.10  & 67.34  &  84.46 & 67.83  &  85.45 \\
		ResNet-34 &  64.56 &  81.16  & 66.14 & 82.39 & 68.20& 84.97 & \textbf{68.66} & \textbf{85.47} \\
		\hline
		\parbox{0.4in}{\centering $\Delta$ }& 2.45  &  0.39 &  1.16 &  0.29 &  0.86 & 0.51  & 0.83  & 0.02 \\
		\hline 
	\end{tabular}
	\caption{Comparison of different capacity models on \textit{mini}ImageNet with input images of 84$\times$84. The ResNet-34 model is a variant of~[\hyperref[reference:He_2016_CVPR-supp]{\textcolor{green}{S-8}}] which is described in Sec. S1-2.   $^{\dagger}$ Reproduced  with our setting.}\label{tab:higher_miniImageNet}
\end{table}

\begin{table}
	\centering
	\footnotesize
	\setlength\tabcolsep{1.2pt}
	\renewcommand\arraystretch{1.3}
	\begin{tabular}{l|c|c|c|c|c|c|c|c}
		\hline
		\multirow{2}*{Method}   & \multicolumn{2}{c|}{ProtoNet$^{\dagger}$}  &  \multicolumn{2}{c|}{Good-Embed$^{\dagger}$} &  \multicolumn{2}{c|}{Meta DeepBDC} &  \multicolumn{2}{c}{STL DeepBDC} \\
		\cline{2-9}
		& 1-shot & 5-shot & 1-shot & 5-shot & 1-shot & 5-shot & 1-shot & 5-shot  \\
		\hline 
		ResNet-18 & 80.90 & 89.81 & 77.92  &    89.94  &  83.55 & 93.82  &  84.01 &  94.02\\
		ResNet-34 &  80.58 &  90.11  & 79.33 & 90.10 & \textbf{85.25} & 94.31 & 84.69 & \textbf{94.33} \\
		\hline
		\parbox{0.4in}{\centering $\Delta$ }& -0.32  &  0.30 &  1.41 &  0.16 &  1.70 & 0.49  & 0.68  & 0.31\\
		\hline 
	\end{tabular}
	\caption{Comparsion of different capacity models  on CUB with input images of 224$\times$224. Here we use  ResNet-34 proposed in~[\hyperref[reference:He_2016_CVPR-supp]{\textcolor{green}{S-8}}].  $^{\dagger}$ Reproduced  with our setting.}\label{tab:higher_CUB}
\end{table}

\subsection*{S3-2~ Effect of Higher Capacity Models}

To evaluate the effect of higher  capacity models, we conduct experiments on CUB using the ResNet-34 with 224$\times$224 input images, and on \textit{mini}ImageNet using the variant of ResNet-34 with  input images of 84$\times$84. We compare our Meta DeepBDC and STL DeepBDC with their respective baselines, i.e., ProtoNet and Good-Embed. 

The comparison on  \textit{mini}ImageNet  is shown in Tab.~\ref{tab:higher_miniImageNet}. It can be seen  that, for every method with either setting, the accuracy obtained by using  high-capacity ResNet-34 is higher than that using ResNet-12. Among them, the improvements of ProtoNet and Good-Embed for 1-shot task are significant (over 1 percentage points). Despite the improvements, our Meta DeepBDC and STL DeepBDC outperform their corresponding baselines by large margins.  
According to Tab.~\ref{tab:higher_CUB} which presents results on CUB, we can see that overall all methods improve when using higher-capacity ResNet-34, while the improvements of ProtoNet are not significant. Again, we observe that our methods are significantly superior  to their baselines.

\subsection*{S3-3~ Effect of Feature Number on DeepBDC}

To obtain more convolutional features,  CTX~[\hyperref[reference:CTX-supp]{\textcolor{green}{S-3}}] removes  the last  down-sampling  in the backbone networks, while  ADM~[\hyperref[reference:ADM-supp]{\textcolor{green}{S-12}}] removes the last two. Similar to them, we also remove down-sampling. On \textit{mini}ImageNet,  we evaluate the effect of down-sampling on   our DeepBDC.  As the input resolution is $84\times 84$ for ResNet-12,  the spatial size of feature maps outputted by the original backbone is  $5\times 5$ and thus we have a total of  25 features; the spatial sizes  become $10\times 10$ and $21\times21$ if the last down-sampling  and the last two are eliminated, respectively.

Tab.~\ref{tab:downsample} summarize the results. We first notice that   variation of feature number has minor effect on  our STL DeepBDC for either  1-shot or 5-shot  task. Regarding our Meta DeepBDC, it can be seen that for both 1-shot and 5-shot tasks,  the accuracy increases slightly when  the number of features is 100, but then decreases when provided with 441 features.  At last, we note that ProtoNet and Good-Embed achieves individual best results when the feature number is 100 and 25, respectively.  Throughout the main paper, we report results with removal of the last down-sampling. 

\begin{table}[t]
	\centering
	\footnotesize
	\setlength\tabcolsep{2.5pt}
	\renewcommand\arraystretch{1.3}
	\begin{tabular}{l|c|c|c|c|c|c}
		\hline
		\multirow{2}*{Method}   & \multicolumn{2}{c|}{5$\times$5}  &  \multicolumn{2}{c|}{10$\times$10} & \multicolumn{2}{c}{21$\times$21}\\
		\cline{2-7}
		& 1-shot & 5-shot & 1-shot & 5-shot & 1-shot & 5-shot \\
		\hline
		ProtoNet~[\hyperref[reference:snell2017prototypical-supp]{\textcolor{green}{S-20}}]~$^{\dagger}$   & 61.81& 79.62& 62.11 & 80.77 &61.45 &80.00\\
		Good-Embed~[\hyperref[reference:tian2020rethinking-supp]{\textcolor{green}{S-21}}]~$^{\dagger}$    & 65.73 & 83.08 & 64.98 & 82.10 & 64.68 & 81.85\\
		\hline
		Meta DeepBDC   & 66.74 & 83.83 & 67.34 & 84.46 & 66.83 & 84.20 \\
		STL DeepBDC   & 67.76 & 85.39 & 67.83 & 85.45 & 67.44 & 85.44\\
		\hline
		
	\end{tabular}
	\caption{Accuracy (\%) against   number of features on \textit{mini}ImageNet for 5-way classification.  $^{\dagger}$ Reproduced  with our setting.}\label{tab:downsample}
\end{table}

\begin{table}[t]
	\centering
	\footnotesize
	\setlength\tabcolsep{3.5pt}
	\renewcommand\arraystretch{1.3}
	\begin{tabular}{c|c|c|c|c|c|c}
		\hline
		\multirow{2}*{Method}   & \multicolumn{2}{c|}{\cellcolor{gray}{Meta-training}}  &  \multicolumn{2}{c|}{Meta-testing} & \multicolumn{2}{c}{Accuracy}\\
		
		\hhline{|~|------|}
		& \cellcolor{gray}{1-shot} & \cellcolor{gray}{5-shot} & 1-shot & 5-shot & 1-shot & 5-shot \\
		\hline
		ProtoNet~[\hyperref[reference:snell2017prototypical-supp]{\textcolor{green}{S-20}}]~$^{\dagger}$  & \cellcolor{gray}{304} & \cellcolor{gray}{365} &  115  & 143 & 62.11 & 80.77\\
		ADM~[\hyperref[reference:ADM-supp]{\textcolor{green}{S-12}}]~$^{\dagger}$   & \cellcolor{gray}{908}  & \cellcolor{gray}{967} & 199 & 221 & 65.87  & 82.05\\
		CovNet~[\hyperref[reference:Wertheimer2019-supp]{\textcolor{green}{S-24}}]~$^{\dagger}$   & \cellcolor{gray}{310} & \cellcolor{gray}{374} & 120 & 144 & 64.59 & 82.02\\
		DeepEMD~[\hyperref[reference:zhang2020deepemd-supp]{\textcolor{green}{S-27}}]   & \cellcolor{gray}{$>$80K} & \cellcolor{gray}{$>$$10^{6}$} & 457 & 12,617 & 65.91 & 82.41\\
		\hline
		Meta DeepBDC   & \cellcolor{gray}{505} & \cellcolor{gray}{623} &  161 & 198 & 67.34 & 84.46\\
		\hhline{|~|------|}
		STL DeepBDC   &  \multicolumn{2}{c|}{\cellcolor{gray}{--}}   & 184 & 245 & 67.83 & 85.45\\
		\hline
		
	\end{tabular}
	\caption{\label{tab:latency} Comparison of latency (ms) for 5-way classification on \textit{mini}ImageNet. $^{\dagger}$ Reproduced  with our setting. }
\end{table}

\subsection*{S3-4~ Comparison of Latency of Meta-training Task}

In the main paper, we compare the  latency of meta-testing task. Here, we give additional comparison for  meta-training task, which  is also important in practice. The latency is measured  on \textit{mini}ImageNet with the backbone of ResNet-12.  Following the setting of DeepEMD~[\hyperref[reference:zhang2020deepemd-supp]{\textcolor{green}{S-27}}], we adopt  QPTH solver~[\hyperref[reference:OptNet-2017-supp]{\textcolor{green}{S-2}}] in meta-training and OpenCV solver in meta-testing,  using the code released by the ~authors\footnote{\href{https://github.com/icoz69/DeepEMD}{https://github.com/icoz69/DeepEMD}}.

Tab.~\ref{tab:latency} shows comparison results for 5-way classification; for reference, we also include latency of meta-testing task and recognition accuracies which have been discussed in the primary paper.    As regards  the meta-training, we find that ProtoNet and CovNet are fastest, while our Meta DeepBDC is somewhat  slower than them. Though the meta-testing speed of ADM is comparable to that of our method, its meta-training latency is much larger than ours; the reason is that  backpropagation of ADM involves GPU-unfriendly matrix inversions. Notably, DeepEMD is at least 80 times slower for 1-shot and 1000 times slower for 5-shot than the other methods; we mention that FRN also observes the big latency of DeepEMD~[\hyperref[reference:Wertheimer_2021_CVPR-supp]{\textcolor{green}{S-25}}].

\begin{figure}[t]
	\centering
	\begin{minipage}[b]{1\linewidth}
		\centering
		\includegraphics[width=0.95\textwidth]{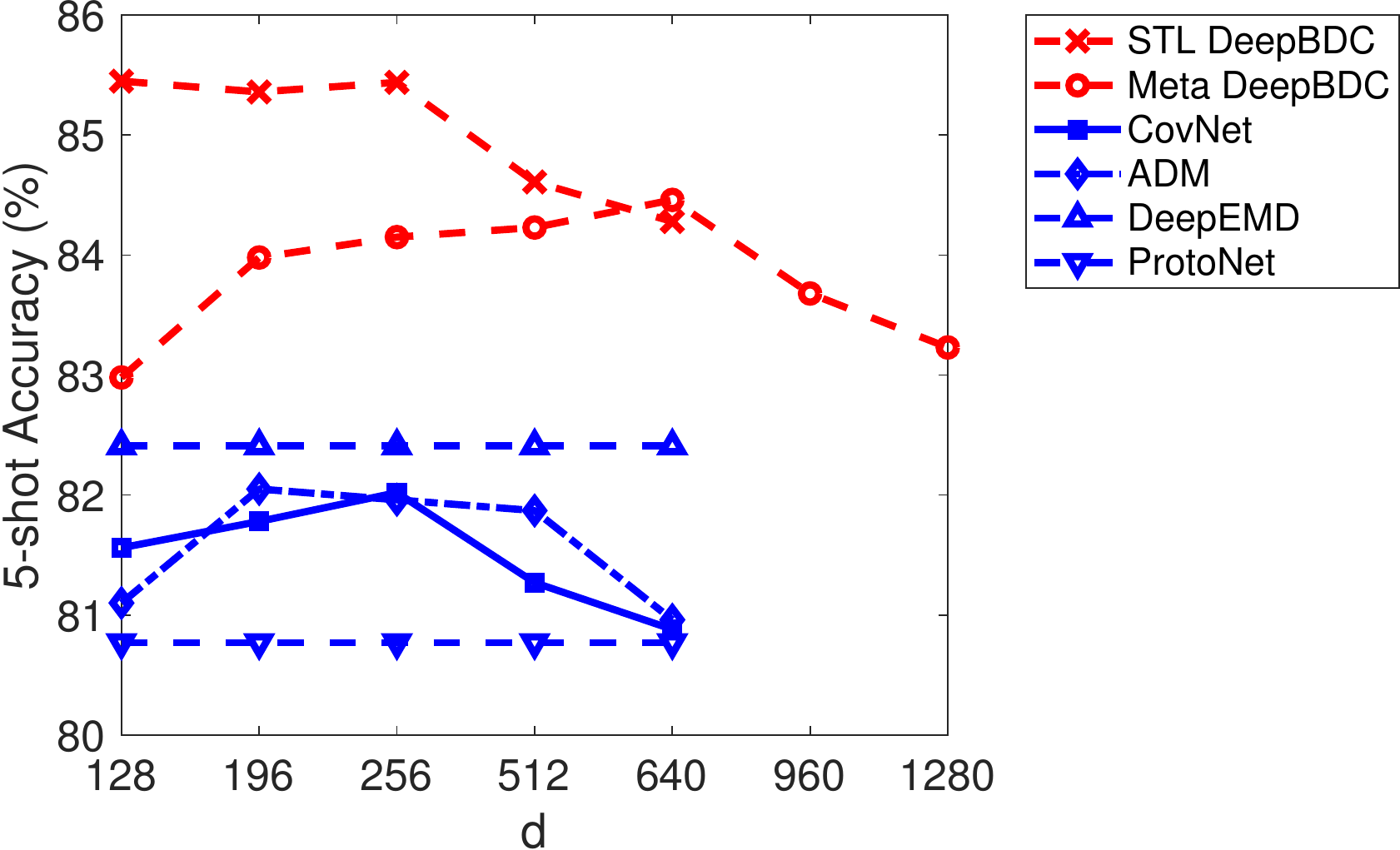}
	\end{minipage}
	\caption{Accuracy as a function of channel number $d$ for 5-way 5-shot classification on \textit{mini}ImageNet.}\label{figure:dim_compare}
\end{figure}

\begin{figure*}[t!]
	\centering
	\begin{subfigure}[b]{1\textwidth}
		\centering
		\includegraphics[height=2.1in]{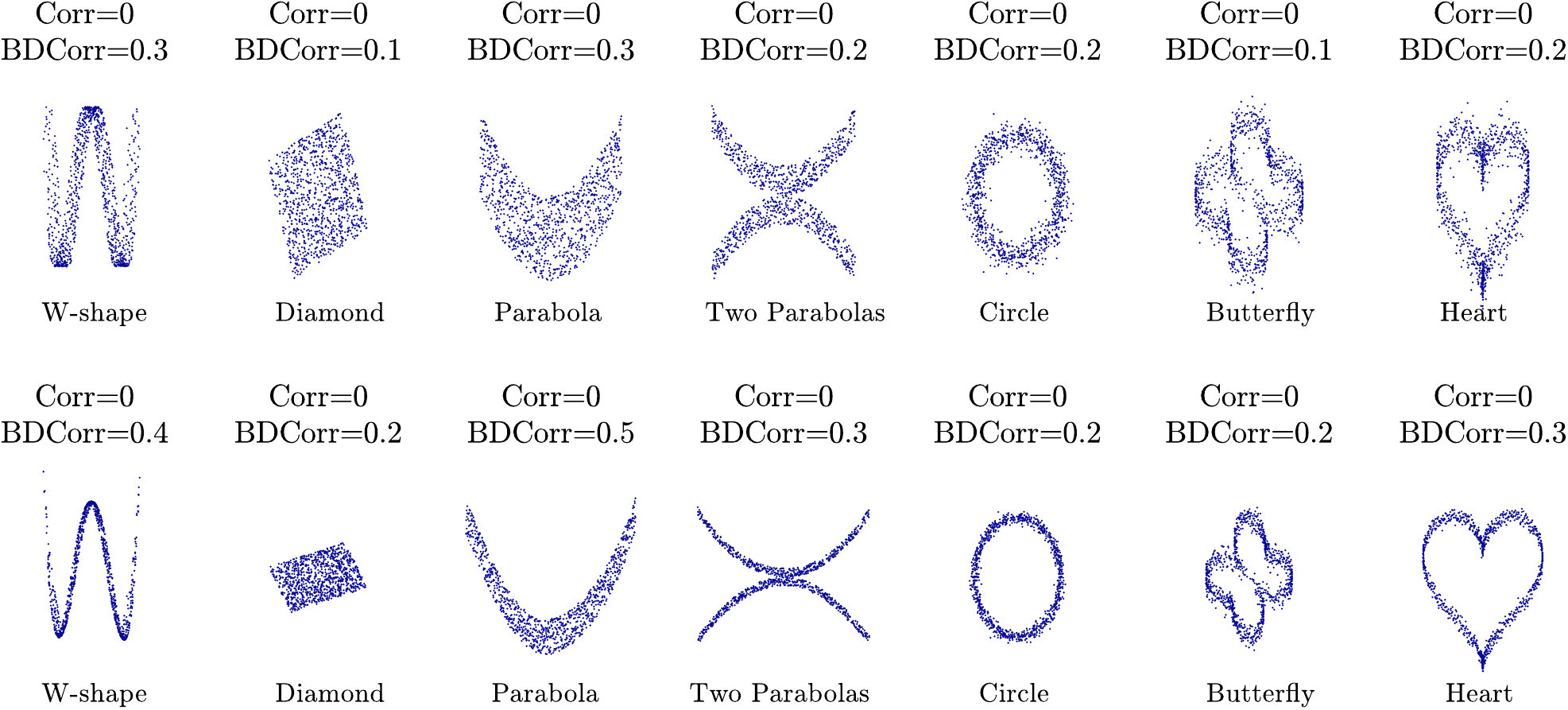}
		\vspace{6pt}
		\caption{Non-linear relations.}
		\label{subfigure:nonlinear}
		\vspace{8pt}
	\end{subfigure}%
	
	\begin{subfigure}[b]{1\textwidth}
		\centering
		\includegraphics[height=2.0in]{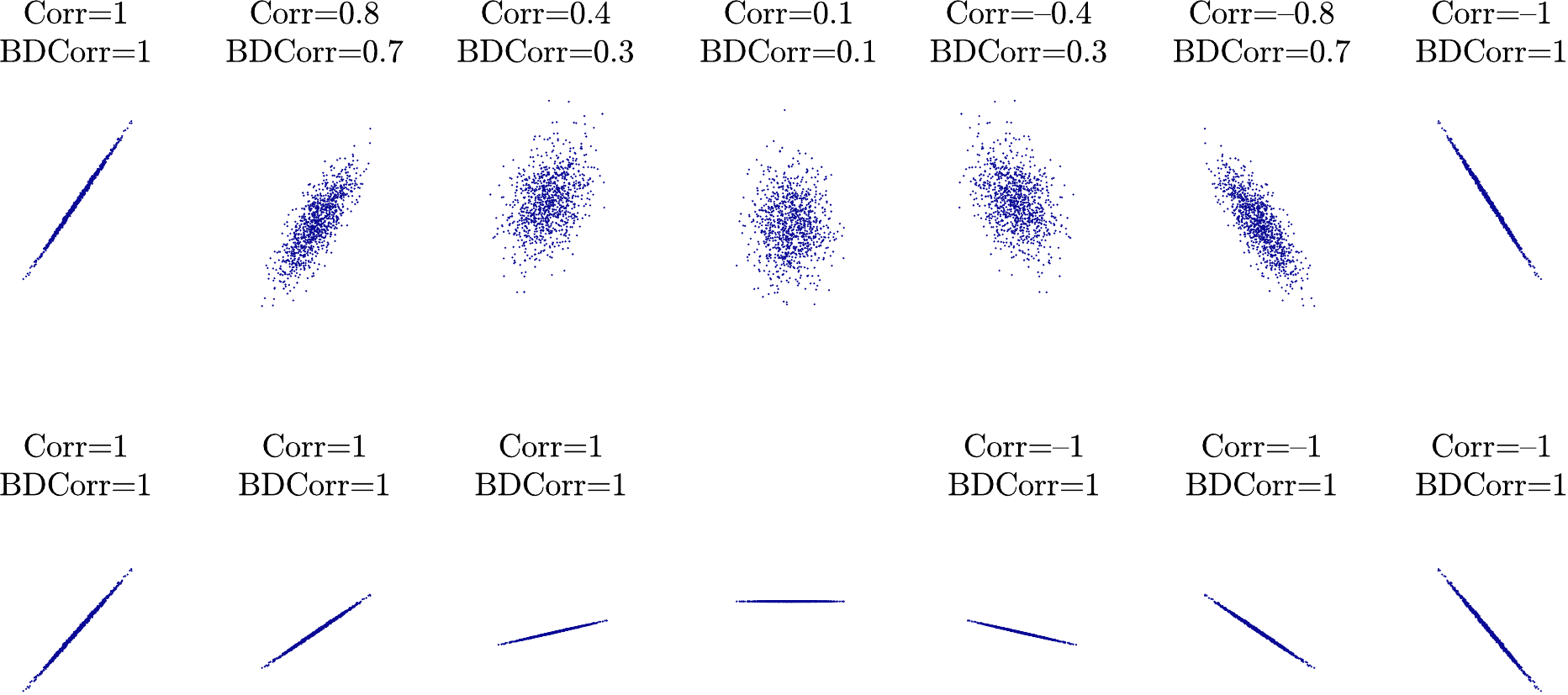}
		\caption{Linear relations.}
		\label{subfigure:linear}
	\end{subfigure}
	\caption{Comparison of non-linear (a) and linear (b) relation modeling   between  classical Correlation ($\mathrm{Corr}$) and Brownian Distance Correlation ($\mathrm{BDCorr}$). This illustration was inspired by~\href{https://commons.wikimedia.org/wiki/File:Distance_Correlation_Examples.svg}{D. Boigelot}.}
	\label{fig:characterize}
\end{figure*}

\subsection*{S3-5~ Effect of Channel Number on  DeepBDC and the Counterparts} \label{subsection: effect of d}

As described in the main paper, both of CovNet and ADM  need to estimate second moments, leading to quadratic increase of representations in channel number $d$. Therefore,  for fair comparison,  we also add a $1\times 1$ convolution for them to reduce the number of channels. Dimension reduction is hurtful for ProtoNet and DeepEMD, so for them we leave the original channel as it is.

Fig.~\ref{figure:dim_compare} plots the curves of accuracies as a function of $d$. In light of   the curves, we can draw several conclusions as follows. (1)  The channel number $d$ has  non-trivial effect on ADM  and  CovNet. The accuracies (\%) of ADM  and CovNet reach the highest values when $d=196$ (82.05) and $d=256$  (82.02), respectively. Their  accuracies   drop gradually  when $d$ becomes  larger, and when $d=640$, they achieve accuracies  only slightly higher than   ProtoNet.    (2)  Across all values of $d$, both instantiations  of our DeepBDC clearly perform better than   the competing methods. 

We mention that our re-implementation non-trivially improves  performance of CovNet and ADM, providing fair and competitive baselines. Besides, these results show that dimension reduction plays an important role for second moment-based methods.


\section*{S4~ Linear and Non-linear Relation Modeling}

One of the favorable properties of Brownian Distance Covariance (BDC) is the ability to model both linear and non-linear dependency between random variables $X$ and $Y$.   In contrast, traditional covariance can only model linear relations.  To facilitate visual understanding, we  consider five  simulated examples of bivariate distributions~[\hyperref[reference:10.2307/43304574-supp]{\textcolor{green}{S-9}}], i.e., ``W-shape'', ``Diamond'', ``Parabola'', ``Two parabolas'' and ``Circle'', and two examples we developed, i.e., ``Butterfly'' and ``Heart'', respectively. In these examples,  two random variables $X$ and $Y$ have different kinds of  non-linear relationships.  Also, we simulate  seven kinds of linear relations based on  HHG package~\footnote{\href{https://cran.r-project.org/web/packages/HHG/index.html}{https://cran.r-project.org/web/packages/HHG/index.html}}.  For  each set of observation pairs,  we compute the classical  correlation
\begin{align}
	\mathrm{Corr}(X,Y)=\dfrac{\mathrm{cov}(X,Y)}{\sqrt{\mathrm{cov}(X,X)}\sqrt{\mathrm{cov}(Y,Y)}}
\end{align}
and Brownian distance correlation
\begin{align}
	\mathrm{BDCorr}(X,Y)=\dfrac{\mathrm{BDC}(X,Y)}{\sqrt{\mathrm{BDC}(X,X)}\sqrt{\mathrm{BDC}(Y,Y)}}
\end{align}
Here $\mathrm{cov}(X,Y)$ and $\mathrm{BDC}(X,Y)$ respectively denote the covariance and Brownian distance covariance. Naturally, $\mathrm{cov}(X,X)$ and $\mathrm{BDC}(X,X)$   denote variance and Brownian distance variance of $X$, respectively.

Fig.~\ref{fig:characterize} shows the scatter plots of the simulated examples together with the values of correlation and Brownian distance correlation. From Fig.~\ref{subfigure:nonlinear}, we can see  that  for all non-linear  relations  $\mathrm{Corr}(X,Y)=0$,  indicating that  classical correlation  fails to  model such complex relations; on the contrary, Brownian distance correlation can characterize the non-linear dependencies. As shown in Fig.~\ref{subfigure:linear}, compared to correlation, Brownian distance correlation has similar capability to model linear relations, except that it cannot distinguish the orientation  as it is always non-negative; besides, both of them cannot reflect the slope of linear relations.

\makeatletter
\def\@bibitem#1{\item\if@filesw \immediate\write\@auxout
	{\string\bibcite{#1}{S-\the\value{\@listctr}}}\fi\ignorespaces}
\def\@biblabel#1{[S-{#1}]}
\makeatother

\phantomsection
\small{

}

\end{appendices}

\end{document}